\documentclass[letterpaper]{article} 
\usepackage{aaai2026}  
\usepackage{times}  
\usepackage{helvet}  
\usepackage{courier}  
\usepackage[hyphens]{url}  
\usepackage{graphicx} 
\usepackage{subfigure}
\usepackage{natbib}  
\usepackage{caption} 
\usepackage{algorithm}
\usepackage{algorithmic}
\usepackage{multirow}
\usepackage{subcaption}
\usepackage{pifont}

\usepackage{colortbl}
\usepackage{amsmath}
\usepackage{booktabs} 
\usepackage{amssymb} 

\usepackage{newfloat}
\usepackage{listings}
\DeclareCaptionStyle{ruled}{labelfont=normalfont,labelsep=colon,strut=off} 
\floatstyle{ruled}
\newfloat{listing}{tb}{lst}{}
\floatname{listing}{Listing}
\title{Head-Aware KV Cache Compression for Efficient Visual Autoregressive Modeling}
\author{
    Ziran Qin\textsuperscript{\rm 1},
    Youru Lv\textsuperscript{\rm 1},
    Mingbao Lin\textsuperscript{\rm 2}\thanks{Corresponding authors.},
    Hang Guo\textsuperscript{\rm 3},
    Zeren Zhang\textsuperscript{\rm 4},
    Danping Zou\textsuperscript{\rm 1},
    Weiyao Lin\textsuperscript{\rm 1}$^*$}
\affiliations{
    \textsuperscript{\rm 1} Shanghai Jiao Tong University, China\\
    \textsuperscript{\rm 2} Rakuten, Singapore\\
    \textsuperscript{\rm 3} Tsinghua University, China\\
    \textsuperscript{\rm 4} Peking University, China\\

    \{qinziran,mishall0914,dpzou,wylin\}@sjtu.edu.cn, linmb001@outlook.com, cshguo@gmail.com, eric\_zhang@stu.pku.edu.cn
}

\usepackage{bibentry}

\begin{document}

\maketitle

\begin{abstract}
Visual Autoregressive (VAR) models adopt a next-scale prediction paradigm, offering high-quality content generation with substantially fewer decoding steps. However, existing VAR models suffer from significant attention complexity and severe memory overhead due to the accumulation of key-value (KV) caches across scales. In this paper, we tackle this challenge by introducing KV cache compression into the next-scale generation paradigm. We begin with a crucial observation: attention heads in VAR models can be divided into two functionally distinct categories: \textit{Contextual Heads} focus on maintaining semantic consistency, while \textit{Structural Heads} are responsible for preserving spatial coherence. This structural divergence causes existing one-size-fits-all compression methods to perform poorly on VAR models. To address this, we propose \textbf{HACK}, a training-free \textbf{H}ead-\textbf{A}ware KV cache \textbf{C}ompression framewor\textbf{K}. HACK utilizes an offline classification scheme to separate head types, enabling it to apply pattern-specific compression strategies with asymmetric cache budgets for each category. By doing so, HACK effectively constrains the average KV cache length within a fixed budget $B$, reducing the theoretical attention complexity from $\mathcal{O}(n^4)$ to $\mathcal{O}(Bn^2)$.
Extensive experiments on multiple VAR models across text-to-image and class-conditional tasks validate the effectiveness and generalizability of HACK. It achieves up to 70\% KV cache compression without degrading output quality, resulting in memory savings and faster inference. For example, HACK provides a $1.75\times$ memory reduction and a $1.57\times$ speedup on Infinity-8B. 
\end{abstract}

\begin{links}
    \link{Code}{https://github.com/Zr2223/HACK}
\end{links}

\section{Introduction}
Autoregressive (AR) models~\cite{he2024mars,sun2024autoregressive,li2024autoregressive} have demonstrated promising performance in visual generation, rivaling diffusion models~\cite{ho2020denoising,podell2023sdxl,peebles2023scalable,esser2024scaling,chen2024pixart} in both quality and flexibility.
However, conventional AR models follow a ``next-token'' prediction paradigm, requiring long decoding steps that suffer from substantial inference latency. 
Recently, Visual AutoRegressive (VAR) modeling~\cite{tian2024visual,han2024infinity} has emerged, replacing the ``next-token'' prediction with a ``next-scale'' prediction paradigm. By enabling the parallel generation of multiple tokens within a single step, VAR models generate high-quality content in a multi-scale, coarse-to-fine manner within a few steps, effectively balancing generation quality and efficiency. 

\begin{figure}[t]
  \centering
  \includegraphics[width=0.95\linewidth]{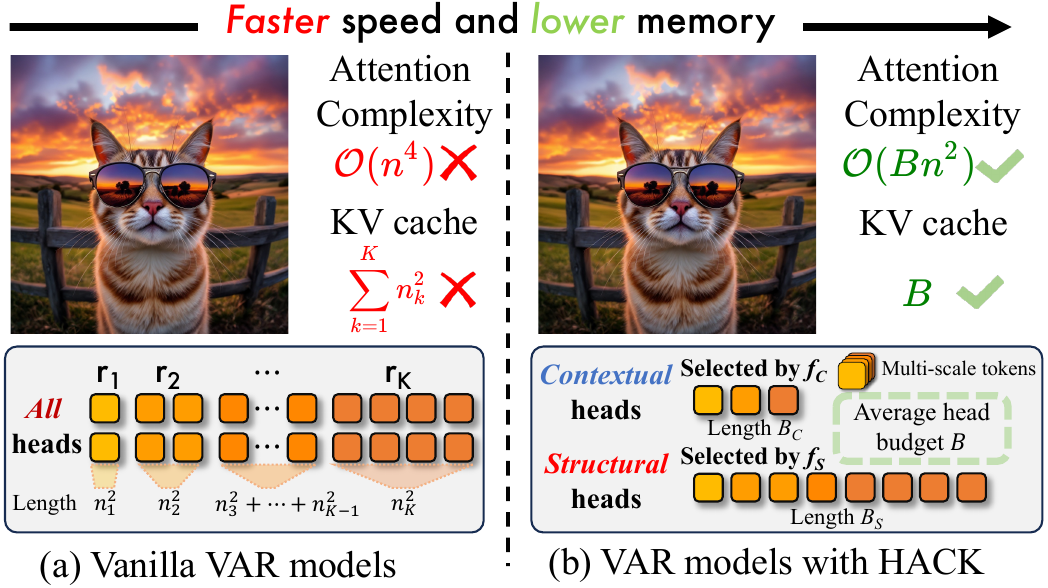}
  \caption{(a) Vanilla VAR models cache all KV pairs across different scales. (b) HACK only preserves proper KV pairs selected by head-aware strategies, effectively hacking down both attention complexity and KV cache length.}
  \label{intro}
\end{figure}

However, VAR models suffer from memory and computational bottlenecks during inference. As shown in Figure~\ref{intro}(a), vanilla VAR indiscriminately caches all key-value (KV) pairs from all previous scales, leading to high attention complexity $\mathcal{O}(n^4)$ and even greater KV cache accumulation than conventional AR models.  
This problem is severe for large-scale content. 
For example, generating $1024\times 1024$ images with Infinity~\cite{han2024infinity} requires processing over 10k tokens across multiple scales, imposing substantial burdens on both attention computation and KV cache storage. This poses a major challenge for practical deployment.

\begin{figure}[!t]
  \centering
  \includegraphics[width=\linewidth]{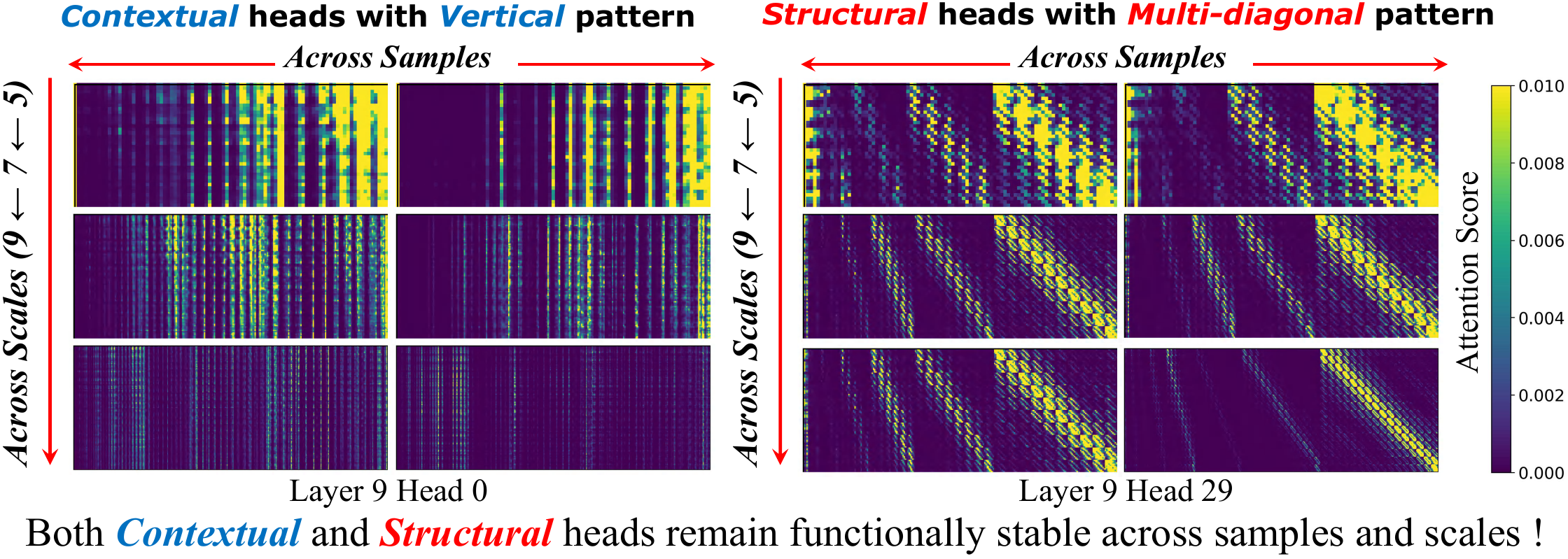}
  \caption{
    \textbf{Attention Patterns of Contextual and Structural Heads.}
    Both Contextual and Structural heads exhibit consistent vertical and multi-diagonal patterns, respectively, across different samples and scales.}
    \label{obser1}
\end{figure}

\begin{figure}[t]
  \centering
  \includegraphics[width=\linewidth]{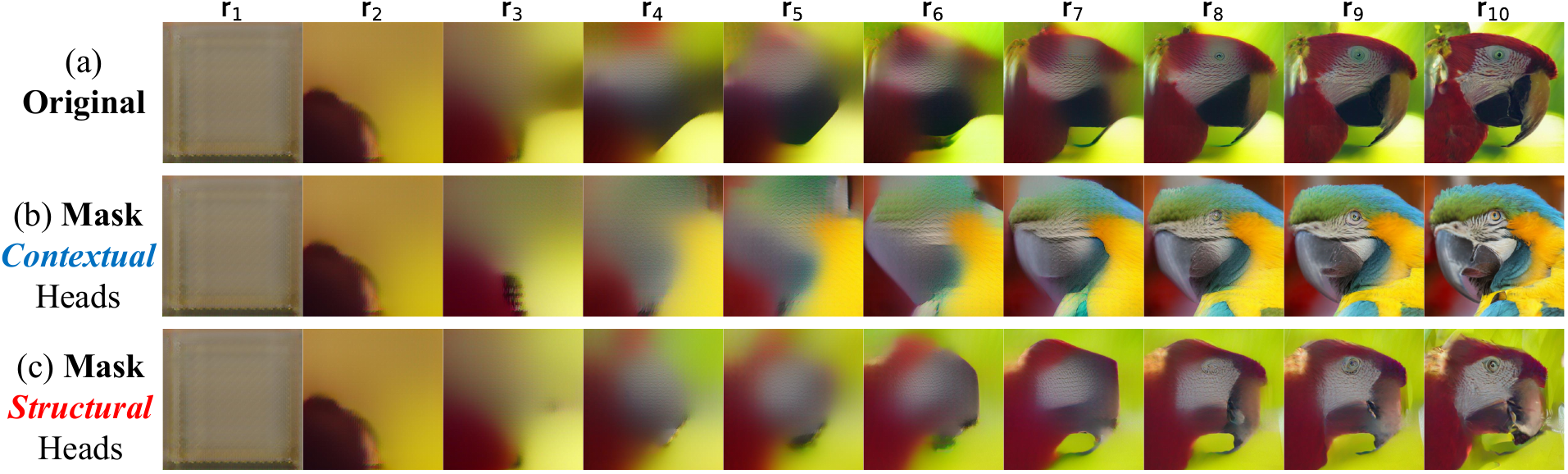}

  \caption{
\textbf{Impact of selective head masking} (10\% of total heads for each type).
Compared with the original generation (a), masking \textit{contextual} heads (b) leads to semantic divergence while maintaining high visual fidelity; in contrast, masking 
\textit{structural} heads (c) preserves global content direction but results in severe structural degradation.
}
\label{maskinghead}
\end{figure}

Inspired by the KV cache optimization in large language models (LLMs)~\cite{achiam2023gpt,anthropic2024claude3,dubey2024llama,liu2024deepseek}, we address these challenges in VAR by compressing unimportant KV pairs before its attention computation. However, existing LLMs-oriented KV compression methods~\cite{zhang2024h2o,li2024snapkv,qin2025cake,wan2024look} fail to generalize to VAR models due to fundamental differences in their attention mechanisms. 
In particular, we observe that attention heads in VAR (shown in Figure~\ref{obser1}) fall into two categories with distinct attention patterns and functional roles. \textit{\textbf{Contextual heads}} focus on a small set of key tokens to maintain global semantic consistency, forming vertical attention patterns.  \textit{\textbf{Structural heads}} attend to spatially adjacent tokens across scales through multi-diagonal attention patterns, preserving the geometric structure of contents.  We further validate their functional roles by selectively masking each type during generation.  In Figure~\ref{maskinghead}, masking contextual heads causes the generated content to semantically diverge, while still maintaining structural integrity and high visual fidelity, whereas masking structural heads preserves the global content direction but causes severe spatial distortions.  
Given the distinct attention characteristics and functional roles of them, existing one-size-fits-all compression strategies are ill-suited for VAR models. This calls for a head-aware KV cache compression approach considering the distinct properties and sensitivities of contextual and structural heads. 

We propose 
\textbf{HACK}, a novel, training-free, head-aware KV cache compression framework for efficient VAR models. HACK first exploits the inherent differences between contextual and structural heads through a robust offline classification based on attention variance.
Then, as shown in Figure~\ref{intro}(b), HACK allocates asymmetric KV budgets and applies pattern-specific compression strategies tailored to each head type. This reduces the attention complexity from $\mathcal{O}(n^4)$ to $\mathcal{O}(Bn^2)$, where $B$ is the average cache length per head. Finally, HACK achieves substantial memory and latency reductions without degrading generation quality.
To the best of our knowledge, HACK is the first study to address KV cache compression in VAR models.

Our contributions include:
(1) We provide an in-depth analysis that identifies and characterizes contextual and structural heads in VAR models, revealing their consistent functional roles and attention patterns, which offers crucial insights for optimizing these models.
(2) We introduce HACK, a head-aware, training-free framework for VAR models that features asymmetric cache budget allocation and pattern-specific KV cache compression strategies suited for both contextual and structural heads.
(3) We prove HACK's efficacy on six different VAR models across multiple tasks, showing substantial memory and latency reductions while maintaining or even improving generation quality.

\section{Related Work}
\textbf{Autoregressive Visual Generation.} AR models~\cite{achiam2023gpt,anthropic2024claude3,dubey2024llama,liu2024deepseek}, originally successful in LLMs, have extended into the visual domain. 
Traditional visual AR approaches~\cite{he2024mars,sun2024autoregressive,li2024autoregressive} rely on next-token prediction, where images are first quantized into discrete tokens (\emph{e.g.}, VQVAE~\cite{van2017neural}, VQGAN~\cite{esser2021taming}) and then decoded autoregressively. While effective, they incur high computation and quantization errors, resulting in lower efficiency.
%
Recent VAR models~\cite{tian2024visual, han2024infinity, tang2024hart, chen2024toward, ren2024m} adopt a next-scale prediction paradigm, enabling multi-scale parallel decoding that improves both quality and speed. They have shown strong performance across text-to-image~\cite{han2024infinity,voronov2024switti}, 3D content~\cite{chen2024sar3d,gao2025mars}, and unified vision-language tasks~\cite{zhuang2025vargpt}. However, their hierarchical design leads to growing attention complexity and KV cache accumulation across scales, resulting in both memory and compute bottlenecks. We address these challenges via VAR-specific KV cache compression that reduces resource cost while preserving generation quality.

\textbf{KV Cache Compression.}
Existing efforts mainly focus on quantization, eviction, and merging to alleviate KV cache overhead in LLMs. Quantization~\cite{liu2024kivi,yue2024wkvquant,kang2024gear,he2024zipcache} reduces cache size by lowering bit precision. 
Eviction methods discard less critical tokens based on attention metrics~\cite{zhang2024h2o, liu2024scissorhands,oren2024transformers,ren2024efficacy,li2024snapkv}, optimizing cache allocation under fixed budgets~\cite{ge2023model,yang2024pyramidinfer,feng2024ada,fu2024not,qin2025cake}. Merging strategies~\cite{zhang2024cam,liu2024minicache,wan2024d2o,wan2024look} alleviate memory constraints by combining redundant KV pairs while preserving context.
However, these methods fall short of VAR models, whose hierarchical attention demands careful preservation of structural and contextual dependencies. We introduce head-aware compression tailored to VAR’s heterogeneous attention patterns, enabling substantial memory and compute reductions without degrading visual quality.

\section{Preliminaries}

\textbf{Visual AutoRegressive Modeling.} VAR extends autoregressive modeling from ``next-token'' to ``next-scale'' prediction. Given an image feature map $\mathbf{f} \in \mathbb{R}^{h \times w \times c}$, VAR quantizes it into $K$ multi-scale token maps $\mathbf{R} = (\mathbf{r}_1, \mathbf{r}_2, \ldots, 
\mathbf{r}_K)$. Each map $\mathbf{r}_k \in [V]^{h_k \times w_k}$ contains $h_k \times w_k$ tokens and the resolutions $\{(h_k, w_k)\}_{k=1}^{K}$ are predefined such that $h_k w_k = a^{2(k-1)}$, where $a > 1$ is a constant scaling factor.
The autoregressive likelihood is:
\begin{equation}
    p(\mathbf{r}_1, \mathbf{r}_2, \ldots, \mathbf{r}_K) = \prod_{k=1}^{K} p(\mathbf{r}_k | \mathbf{r}_1, \mathbf{r}_2, \ldots, \mathbf{r}_{k-1}),
\end{equation}
where the generation of each scale $k$ is conditioned on the tokens $\{\mathbf{r}_1, \mathbf{r}_2, \ldots, \mathbf{r}_{k-1}\}$ from all preceding scales, which serve as a contextual prefix.
During inference,  the entire content can be synthesized coarse-to-fine in just $K$ steps, with all tokens for each scale predicted in parallel.

\textbf{KV Cache in VAR.}
Akin to AR models, VAR improves efficiency by caching KV pairs across generation steps. In a multi-head attention module, each head $h$ has weight matrices $\mathbf{W}_Q^{(h)}, \mathbf{W}_K^{(h)}, \mathbf{W}_V^{(h)} \in \mathbb{R}^{D \times D_h}$, projecting input tokens $\mathbf{X}_k \in \mathbb{R}^{N_k \times D}$ at step $k$ into queries, keys, and values:
\begin{equation}
\small
\mathbf{Q}_k^{(h)} = \mathbf{X}_k \mathbf{W}_Q^{(h)}, \; \mathbf{K}_k^{(h)} = \mathbf{X}_k \mathbf{W}_K^{(h)}, \; \mathbf{V}_k^{(h)} = \mathbf{X}_k \mathbf{W}_V^{(h)},  
\end{equation}
where $N_k = h_k w_k$ denotes the number of tokens at scale $k$. The KV caches are updated independently per head:
\begin{equation}
\small
\label{kvcat}
\mathbf{K}^{(h)} = \text{Concat}(\mathbf{K}^{(h)}, \mathbf{K}_k^{(h)}), \quad \mathbf{V}^{(h)} = \text{Concat}(\mathbf{V}^{(h)}, \mathbf{V}_k^{(h)}),
\end{equation}
with the cache length at step k, denoted as $T_k = \sum_{i=1}^{k} w_i h_i$, growing cumulatively. Attention is then computed as:
\begin{equation}
\small
\mathbf{A}^{(h)}=\text{Softmax}(\mathbf{Q}^{(h)}{\mathbf{K}^{(h)}}^T),\quad
\mathbf{O}^{(h)} = \mathbf{A}^{(h)}\mathbf{V}^{(h)}.
\end{equation}

While caching avoids redundant computations across scales, the accumulation of tokens leads to escalating memory consumption. Also, the attention cost increases as the sequence length grows, causing complexity of $\mathcal{O}(n^4)$, where $n=a^{k-1}$. These dual challenges make efficient KV cache management a vital prerequisite for scalable inference.

\textbf{KV Cache Compression for VAR.}
To address the above memory and attention compute challenges, we introduce a KV cache compression strategy for VAR models. Without compression, the total cache size can become prohibitively large. We quantify this as $B_\text{total} = T_K\cdot H\cdot L$, where $T_K$ is the total number of generated tokens, $H$ and $L$ are the number of attention heads and layers, respectively. Our strategy is to enforce a fixed per-head budget $B$. Whenever the cache length $T_k$ exceeds $B$ at any generation step $k$, compression is applied to the KV cache before the attention computation:
\begin{equation}
\hat{\mathbf{K}}^{(h)}, \hat{\mathbf{V}}^{(h)} = f(\mathbf{K}^{(h)}, \mathbf{V}^{(h)}, B),
\end{equation}
where $\hat{\mathbf{K}}^{(h)}, \hat{\mathbf{V}}^{(h)} \in \mathbb{R}^{B \times D_h}$ are the compressed KV pairs, and $f(\cdot)$ is any compression strategy. 

\begin{figure}[t]
  \centering
      \includegraphics[width=0.45\textwidth]{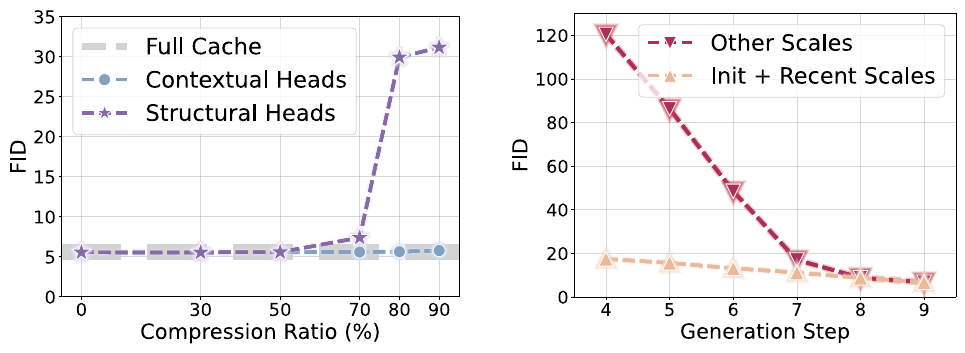}
      \caption{\textbf{Empirical analysis on VAR-d30.} (Left) Compression sensitivity of contextual vs. structural heads. (Right) Comparison of two scale-preserving strategies. Experiments are conducted on 8K ImageNet samples.}
      \label{fig:ober23}
\end{figure}

\begin{figure*}[t]
  \centering
  
  \includegraphics[width=0.95\linewidth]{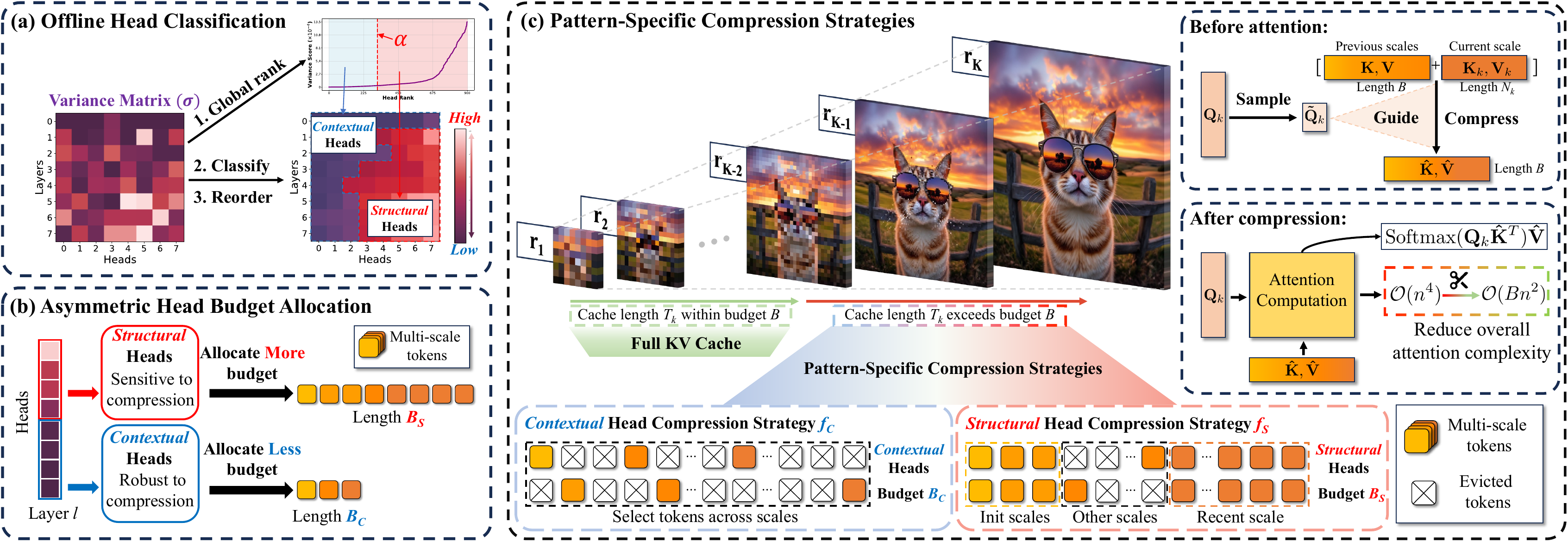}
    \caption{\textbf{Overview of our proposed HACK framework.} 
It consists of (a) offline head classification via attention variance, 
(b) asymmetric head budget allocation based on compression sensitivity, and (c) pattern-specific KV cache compression strategies.}
  \label{fig:overall}
\end{figure*}

\section{Methodology}

\subsection{Motivation}
\label{sec:in-depth}
To guide the design of KV compression for VAR models, we first conduct a series of empirical analyses using VAR-d30. Our observations are summarized as follows:

\textbf{\textit{Observation 1: Stability of attention heads across samples and scales.}}
As shown in Figure\,\ref{obser1}, despite variations in input samples or generation scales, both contextual and structural heads consistently exhibit their respective vertical and multi-diagonal attention patterns. This consistency suggests that attention heads in VAR models play stable, functional roles that are intrinsic to the model and can thus be reliably classified. We provide more attention visualization results in the \textit{supplementary material}.

\textbf{\textit{Observation 2: Divergent compression sensitivity across head types.}} The two head types show markedly different sensitivities to compression. Contextual heads, which focus attention on key tokens to summarize global semantics, are resistant to pruning. In contrast, structural heads, which maintain geometric structure on each scale, are highly sensitive, as excessive compression can disrupt critical spatial information. As shown in Figure\,\ref{fig:ober23} (left), contextual heads maintain high image quality even with 90\% compression ratio, while the performance of structural heads degrades significantly when compression exceeds 50\%. This difference in compression sensitivity motivates our strategy of allocating asymmetric budgets to use resources more effectively.

\textbf{\textit{Observation 3: Initial and recent scales are more important in generation.}}
Drawing inspiration from LLMs, where initial and recent tokens receive higher attention and are crucial for coherence~\cite{xiao2023efficient}, we hypothesized that initial and recent scales play a similar role in VAR models. To verify this, we compared two token preservation strategies: (1) retaining tokens only from the initial two and the most recent scales, versus (2) retaining an equal number of tokens from intermediate scales. As shown in Figure\,\ref{fig:ober23} (right), strategy (1) yields dramatically better generation quality, with FID scores up to $10\times$ lower in early generation steps. This confirms the greater importance of tokens from the initial and recent scales in VAR’s generation.

\subsection{Offline Head Classification via Attention Variance}
\label{4-2}
Motivated by the distinct and consistent functional behaviors of contextual and structural heads, we propose an offline classification method (illustrated in Figure\,\ref{fig:overall}(a)) to distinguish them before deployment. As we analyzed, contextual heads exhibit stable attention patterns across queries, selectively focusing on or disregarding semantic cues, thereby yielding low column-wise variance. In contrast, structural heads attend in a position-sensitive manner, dynamically shifting attention based on spatial configurations, which results in higher variance across queries. Therefore, we capture their divergent behaviors by measuring the attention variance. Specifically, we analyze the attention matrix $\mathbf{A}_K^{(l, h)} \in \mathbb{R}^{N_K \times T_K}$ at the final generation step $K$, which encapsulates interactions from all prior scales. For each head $h$ in layer $l$, we compute the sum of column-wise variances:
\begin{equation}
\sigma_{l,h} = \sum_{j=1}^{T_K} \mathrm{Var}(\mathbf{A}_K^{(l, h)}[:, j]).
\end{equation}

Averaging these values over a small validation set yields a variance matrix $\boldsymbol{\sigma} \in \mathbb{R}^{L \times H}$. We then globally rank all heads according to their variance scores. Notably, the resulting distribution exhibits a long-tailed pattern, with a clear boundary separating low-variance and high-variance regions, corresponding to contextual and structural heads, respectively.
To assign head types, we define a global contextual head ratio $\alpha$, which can be tuned based on the empirical variance distribution to suit different models.

Following classification, we obtain per-layer index sets $\mathbb{H}_C$ and $\mathbb{H}_S$ for contextual and structural heads, respectively, satisfying $\mathbb{H}_C \cup \mathbb{H}_S = \{1, \dots, H\}$ and $\mathbb{H}_C \cap \mathbb{H}_S = \emptyset$. For clarity, we annotate head-specific variables with superscripts $(C)$ and $(S)$; for instance, $\mathbf{K}^{(C)}$ and $\mathbf{V}^{(C)}$ denote the key and value states of contextual heads, while $\mathbf{K}^{(S)}$ and $\mathbf{V}^{(S)}$ correspond to those of structural heads. To facilitate efficient inference, we conceptually reorder the heads within each layer by grouping them according to type.

\subsection{Asymmetric Head Budget Allocation}

Given an average target budget $B$ per head and the contextual head ratio $0 < \alpha < 1$, 
we allocate memory asymmetrically to contextual and structural heads:
%
\begin{equation}
B = \alpha B_C + (1 - \alpha) B_S,
\end{equation}
where $B_C$ and $B_S$ denote the cache budget assigned to each contextual and structural head, respectively.
In practice, we assign substantially smaller budgets to contextual heads (\textit{i.e.}, $B_C \ll B_S$), 
preserving more capacity for structural heads due to their higher sensitivity to compression. It is worth noting that although $\alpha$ defines the global proportion of contextual heads, the actual contextual–structural head ratio varies across layers. This naturally induces a layer-adaptive compression effect, where cache budgets are allocated according to each layer’s specific head composition.

When the cache length $T_k$ for a head group exceeds its assigned budget at any step $k$, 
we invoke dedicated compression strategies \textit{prior to attention computation} 
to limit the size of stored KV pairs. 
This ensures that memory and compute costs remain bounded before executing the attention module.
The compressed KV pairs are computed as:
\begin{equation}
\hat{\mathbf{K}}^{(p)}, \hat{\mathbf{V}}^{(p)} = f_p\left(\{\mathbf{K}^{(h)} | h \in  \mathbb{H}_p\}, \{\mathbf{V}^{(h)} | h \in  \mathbb{H}_p\}, B_p\right),
\end{equation}
where $p \in \{C, S\}$ and $f_p$ denotes the specific compression strategy for contextual or structural heads.

\subsection{Pattern-Specific Compression Strategies}\label{4-3}
We design compression strategies, as shown in Figure\,\ref{fig:overall}(c), that are tailored to the distinct attention patterns of contextual and structural heads, enabling fine-grained and head-aware KV cache compression.

\textbf{Importance Estimation via Efficient Subset Attention.} To make compression efficient, we propose to approximate full attention behavior by observing a small subset of queries. For each head type $p \in \{C, S\}$, we uniformly sample a small subset of queries 
$\mathbf{\tilde{Q}}^{(p)}_k \in \mathbb{R}^{N_{\text{obs}} \times D_h}$ from full $\mathbf{Q}^{(p)}_k \in \mathbb{R}^{N_k \times D_h}$ and compute attention scores over uncompressed key matrix $\mathbf{K}^{(p)} \in \mathbb{R}^{T_k \times D_h}$:
\begin{equation}
\mathbf{\tilde{A}}^{(p)}_k = \text{Softmax}(\mathbf{\tilde{Q}}^{(p)}_k (\mathbf{K}^{(p)})^T).
\end{equation}

These approximated scores act as a computationally cheap proxy for full attention, enabling token selection and compression for both contextual and structural heads without incurring the cost of full attention computation.

\textbf{Contextual Head Compression Strategy $f_C$.} 
Contextual heads typically focus on a small number of semantically salient tokens, resulting in vertically concentrated attention patterns. Based on this property, we directly select KV pairs $\{\hat{\mathbf{K}}^{(C)},\hat{\mathbf{V}}^{(C)}\}$ with the highest cumulative attention scores: 
\begin{equation}
\begin{aligned}
\mathbf{I}^{(C)} &= \text{TOPK}\left(\sum_{i=1}^{N_{\text{obs}}}\mathbf{\tilde{A}}^{(C)}_k[i,:], B_C \right), \\
\hat{\mathbf{K}}^{(C)} &= \mathbf{K}^{(C)}[\mathbf{I}^{(C)},:], \quad
\hat{\mathbf{V}}^{(C)} = \mathbf{V}^{(C)}[\mathbf{I}^{(C)},:],
\end{aligned}
\end{equation}
where $\text{TOPK}$ selects the indices of the top-$K$ tokens with highest scores. 
To mitigate potential semantic loss from aggressive pruning at the final generation step, we apply a merging operation following LOOK-M~\cite{wan2024look}. Discarded KV tokens in the last generation step are softly integrated into their nearest retained counterparts through similarity-weighted averaging. Merging is performed only once at the final scale to balance efficiency and performance.

\textbf{Structural Head Compression Strategy $f_S$.}
Unlike contextual heads, structural heads exhibit position- and scale-dependent patterns, making them sensitive to spatial structures and positional continuity. To preserve structural coherence, our scale-aware compression strategy prioritizes tokens from the crucial initial and recent scales. Specifically, we retain $N_{\text{init}}$ tokens from initial generation stages and $N_k$ tokens from the latest scale, using the remaining budget $M$ to select intermediate-scale tokens with the highest cumulative attention scores via subset attention.
The conserved KV caches $\{\hat{\mathbf{K}}^{(S)}$, $\hat{\mathbf{V}}^{(S)}\}$ in structural heads are given by:
\begin{equation}
\small
\begin{aligned}
\mathbf{I}^{(S)} &= \text{TOPK}\left(\sum_{i=1}^{N_{\text{obs}}}\mathbf{\tilde{A}}^{(S)}_k[i,:],\, M\right), \quad M = B_S - N_{\text{init}} - N_k, \\
\hat{\mathbf{K}}^{(S)} &= \text{Concat}\left(\mathbf{K}^{(S)}[:N_{\text{init}},:],\ \mathbf{K}^{(S)}[\mathbf{I}^{(S)},:],\ \mathbf{K}^{(S)}[-N_k:,:]\right), \\
\hat{\mathbf{V}}^{(S)} &= \text{Concat}\left(\mathbf{V}^{(S)}[:N_{\text{init}},:],\ \mathbf{V}^{(S)}[\mathbf{I}^{(S)},:],\ \mathbf{V}^{(S)}[-N_k:,:]\right).
\end{aligned}
\end{equation}

This preserves long-range dependencies and recent local context, while adaptively selecting intermediate tokens to maintain spatial continuity with minimal redundancy.

\subsection{Complexity Analysis}

Compared to vanilla VAR, which incurs a total attention complexity of $\mathcal{O}(n^4)$ due to cumulative KV cache growth (see \textit{supplementary material} for derivation), HACK reduces this cost to $\mathcal{O}(B  n^2)$ by enforcing an average per-head cache budget $B$. At generation step $k$, the attention complexity is upper-bounded by $N_k \cdot \min(T_k, B)$, where $N_k$ is the number of tokens at step \( k \) and \( T_k \) is the cumulative cache length up to step \( k \). Summing over all scales yields:
\begin{equation}
\small
\sum_{k=1}^{K} N_k \cdot \min(T_k, B) 
\leq B \sum_{k=1}^{K} a^{2(k - 1)} 
\sim \mathcal{O}(B n^2).
\end{equation}

The additional overhead in HACK arises from KV cache compression, triggered only when $T_k > B$. Let $k_s$ denote the first step where compression is applied. The main cost comes from subset attention, where $N_{\text{obs}} \ll N_k$ sampled queries estimate token importance:
\begin{equation}
\small
\sum_{k = k_s}^{K} N_{\text{obs}} \cdot (B + N_k) 
= N_{\text{obs}} \sum_{k = k_s}^{K} (B + a^{2(k-1)}) 
\sim \mathcal{O}(N_{\text{obs}} n^2).
\end{equation}

\section{Experimentation}\label{5}
\subsection{Experiment Settings}\label{5-1}
\textbf{Evaluation Models.}
We evaluate multiple next-scale generation models across diverse generative tasks: Infinity-2B, Infinity-8B~\cite{han2024infinity} and Hart~\cite{tang2024hart} for text-to-image generation; VAR-d30~\cite{tian2024visual} for class-conditional generation. 

\textbf{Evaluation Benchmarks and Metrics.} 
For text-to-image generation, we adopt four benchmarks: GenEval~\cite{ghosh2023geneval} for assessing high-level semantic alignment; ImageReward (IR) ~\cite{xu2023imagereward}, HPSv2.1 (HPS)~\cite{wu2023human}, and MJHQ30k (evaluated using FID and CLIP)~\cite{li2024playground} for perceptual quality.
For class-conditional generation, we report FID, Inception Score (IS), Precision, and Recall evaluated on ImageNet-1K~\cite{deng2009imagenet} using 50 samples per class.


\textbf{Baseline Methods.}
We compare against two categories of KV cache compression methods: 
(1) \textit{Eviction-based}, including StreamingLLM~\cite{xiao2023efficient} (position-based), and H2O~\cite{zhang2024h2o}, SnapKV~\cite{li2024snapkv}, and CAKE~\cite{qin2025cake} (attention-based); 
(2) \textit{Merging-based}, including LOOK-M~\cite{wan2024look} and Meda~\cite{wan2025meda}, which merge evicted KV pairs.


\textbf{Implementation Details.}
%
The compression ratio for evaluation is defined by the average per-head budget $B$ as $\rho = 1 - \frac{B}{T_K}$. Additional HACK implementation details for different models are provided in the \textit{supplementary material}.

\begin{table}[!t]
\small
\centering
\begin{tabular}{lccccc}
\toprule
Method  &GenEval $\uparrow$ & HPS$\uparrow$ & IR$\uparrow$ &FID$\downarrow$& CLIP$\uparrow$  \\

\midrule
Infinity-2B &0.68 &30.49 &0.946 &10.34&27.52 \\
+Streaming &0.68&29.76&0.901&11.20&27.54 \\
+H2O &0.68&29.60&0.910&10.68 &27.57 \\
+SnapKV &0.68&29.60&0.904&10.60 &27.56 \\
+LOOK-M &0.67&28.73&0.864&11.14&27.59\\
+CAKE &0.68 & 29.46&0.906&10.59 &27.56\\
+MEDA &0.67 &28.70 &0.867&11.14&27.59\\
+HACK &\textbf{0.68}&\textbf{30.18}&\textbf{0.933}& \textbf{10.56}&\textbf{27.62}\\

\midrule
Infinity-8B &0.81 &30.99 &1.049& 8.75& 28.73\\
+Streaming&0.81&30.52&1.016&8.98&29.05\\
+H2O &0.81&30.53&1.020&9.04&29.02\\
+SnapKV &0.81&30.21&1.015&9.45&29.06\\
+LOOK-M &0.81&29.99&0.994&10.84&29.04\\
+CAKE&0.81 &30.04 &1.002 &9.80 &29.05\\
+MEDA &0.79&29.57 &0.954&11.56&29.01\\
+HACK &\textbf{0.82}&\textbf{30.69}&\textbf{1.043}&\textbf{8.62}&\textbf{29.08}\\

\midrule
Hart&0.50 &28.75&0.658&10.70 &27.69\\
+Streaming&0.48&25.57&0.458&11.16 &27.38\\
+H2O&0.48&25.69&0.475&11.13 &27.32\\
+SnapKV &0.47&25.62&0.469&11.33&27.33\\
+LOOK-M &0.37&20.83&0.140&25.64&25.85\\
+CAKE&0.46&25.89&0.458&12.88&27.21\\
+MEDA&0.36&20.57&0.117&28.37 &25.71\\
+HACK &\textbf{0.50}&\textbf{28.36}&\textbf{0.658}&\textbf{8.58}&\textbf{27.76}\\

\bottomrule
\end{tabular}
\caption{Quantitative comparison of text-to-image generation on different models with compression ratio $\rho=70\%$. 
}
\label{infinityresults}
\end{table}

\subsection{Main Results}
\textbf{Text-to-Image Generation.}
\textit{Quantitative Results:} 
%
%
Table\,\ref{infinityresults} compares HACK with six baseline methods on Infinity (2B/8B) and Hart models at compression ratio $\rho=70\%$. Compared to baseline methods, which suffer performance degradation in perceptual quality, HACK achieves lossless performance across all evaluation metrics and models, demonstrating strong effectiveness and generalizability. Notably, HACK even surpasses the full KV cache in some cases, underscoring significant redundancy in both attention computation and KV cache in vanilla next-scale generation. 

%
\begin{figure}[!t]
  \centering
  \includegraphics[width=0.87\linewidth]{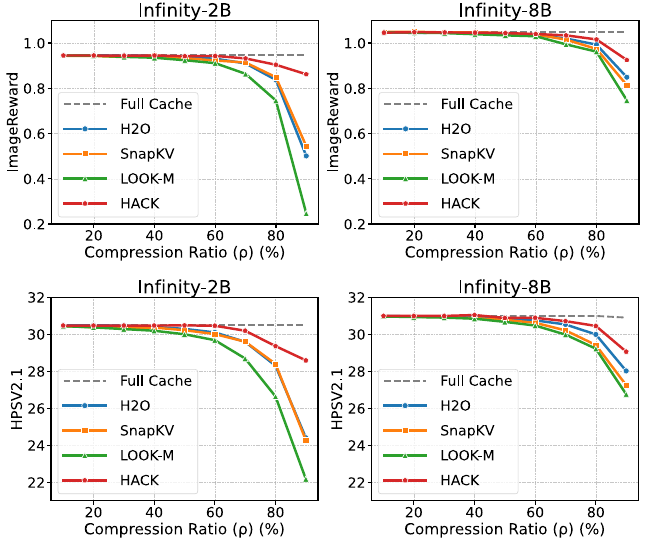}
  \caption{Performance comparison of different compression methods across varying compression ratios measured by ImageReward and HPSv2.1 on Infinity-2B and Infinity-8B.}
\label{sparsityratio}
\end{figure}
\begin{figure*}[!t]
  \centering
  \begin{minipage}[t][6cm][t]{0.65\linewidth}
    \centering   \includegraphics[width=\linewidth, height=5cm, keepaspectratio]{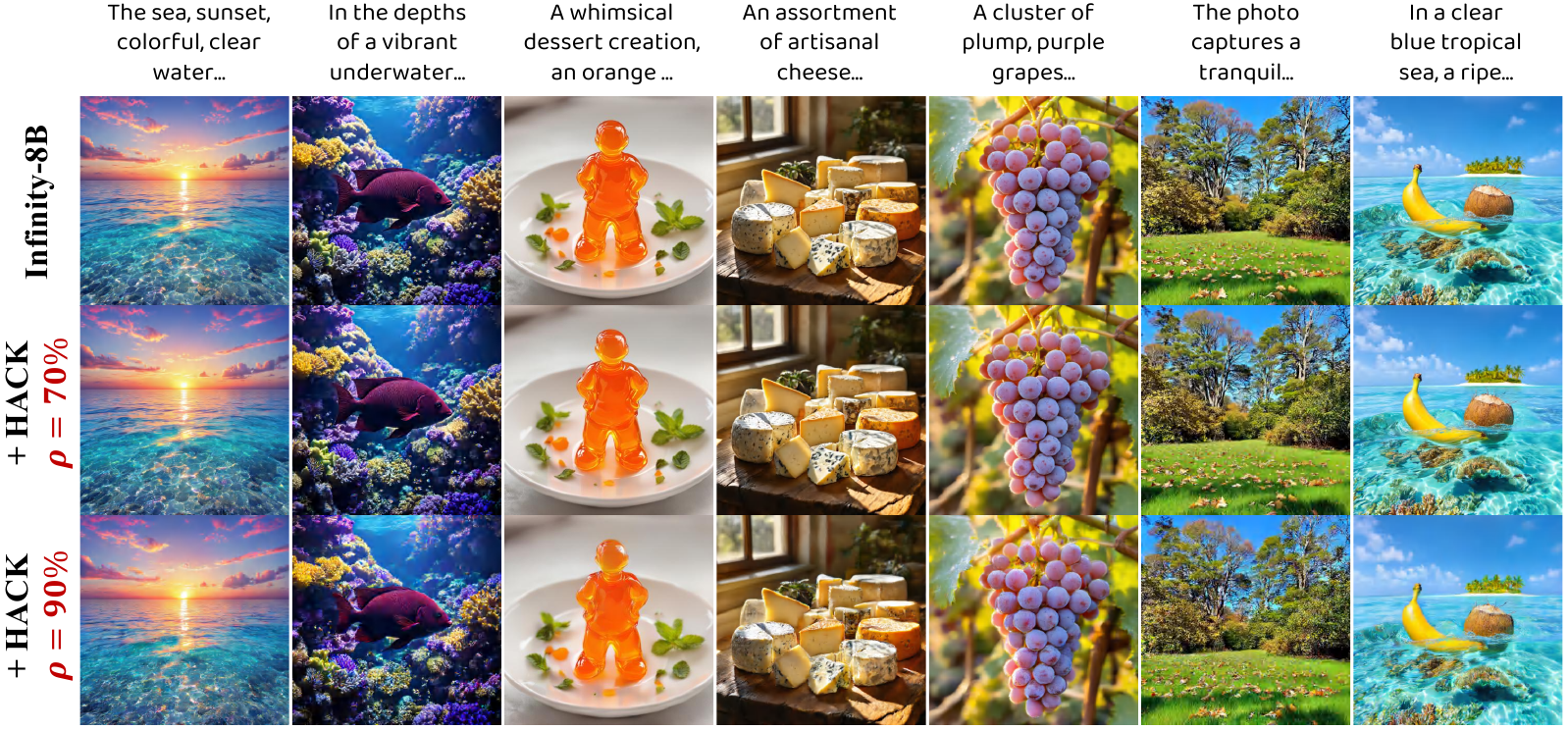} 
\caption{Qualitative results of text-to-image generation by HACK on Infinity-8B. HACK can preserve visual fidelity even under an extreme compression ratio.}

    \label{infinityquan}
  \end{minipage}
  \hfill
  \begin{minipage}[t][6cm][t]{0.34\linewidth}
    \centering  \includegraphics[width=\linewidth, height=5cm, keepaspectratio]{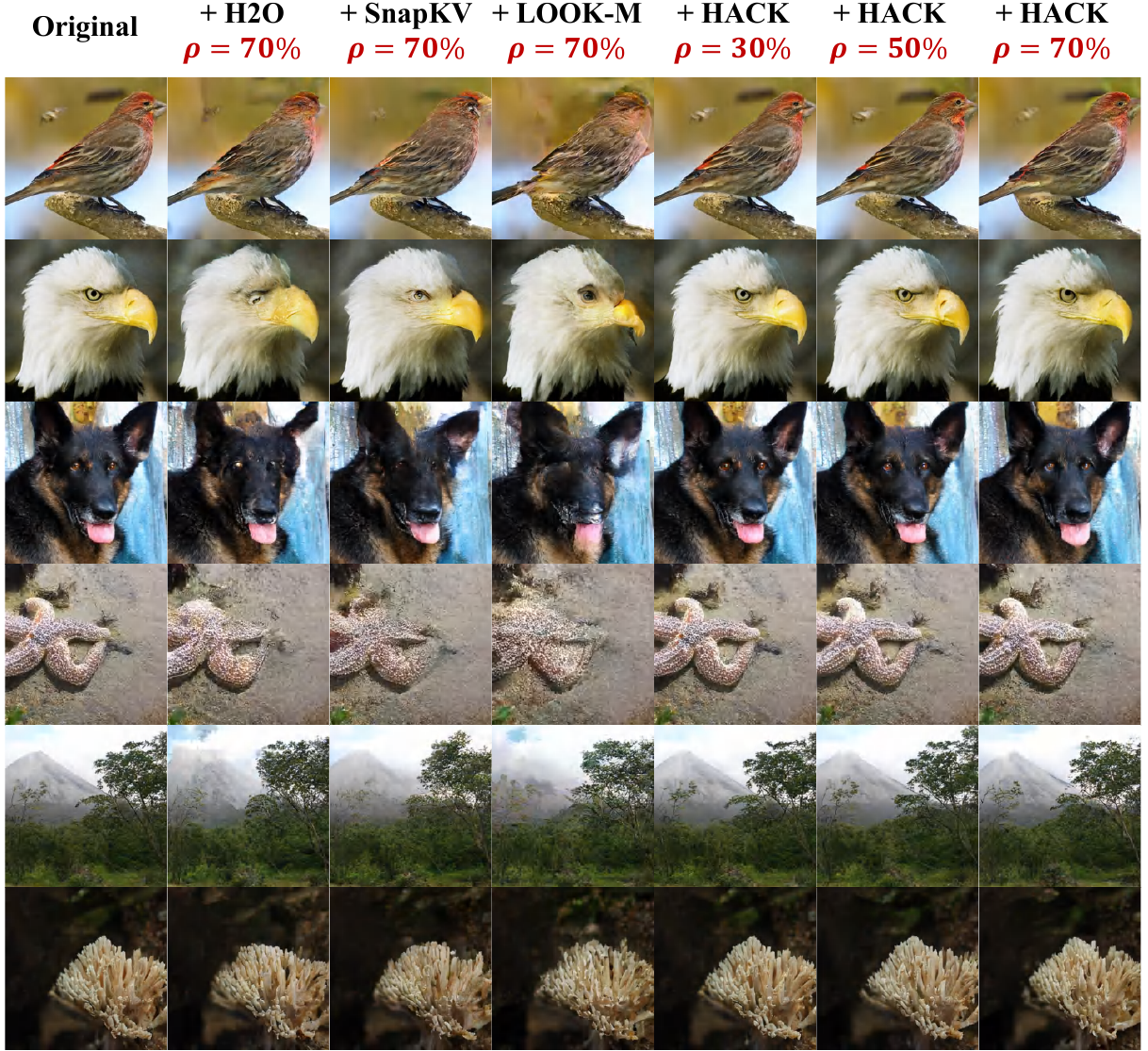} 
\caption{Qualitative comparison of class-conditional image generation on VAR-d30.}
    \label{varquan}
   \end{minipage}
\end{figure*}

\textit{Influence of Compression Ratios:} 
Figure\,\ref{sparsityratio} shows HACK's qualitative results at varying compression ratios on Infinity-2B/8B. While baselines struggle to maintain perceptual quality, resulting in substantial performance drops at extreme compression levels ($\rho>70\%$), HACK consistently preserves robust image quality.

\textit{Qualitative Results:}
We further present qualitative examples from HACK on Infinity-8B in Figure\,\ref{infinityquan}. HACK effectively maintains pixel-level fidelity, scene layout, object structure, and semantic coherence with original outputs. Even at extreme compression ($\rho=90\%$), generated images remain visually faithful, confirming HACK's resilience.

\begin{table}[!t]
\small
\centering

\begin{tabular}{lccccc}
\toprule
Method &$\rho$& FID$\downarrow$ & IS$\uparrow$ & Precision$\uparrow$ & Recall$\uparrow$ \\
\midrule

VAR-d30 &0\% & 1.96  & 302.23 & 0.81 & 0.60 \\
\midrule
+Streaming &30\%&2.04 & 297.09& 0.80& 0.61\\
+H2O &30\%& 2.10  & 295.01 & 0.81 & 0.60 \\
+SnapKV &30\%& 2.13 & 294.85 & 0.81 & 0.60 \\
+LOOK-M &30\%&3.32 & 255.36 & 0.76 & 0.61 \\

+HACK&30\%& \textbf{1.97} & \textbf{299.98} & 0.81 & 0.61 \\
\midrule

+Streaming &50\%&2.36 & 281.30& 0.79& 0.61\\
+H2O &50\% & 3.04 & 262.68 & 0.77 & 0.62 \\
+SnapKV&50\% & 3.09 & 261.63 & 0.77 & 0.62 \\
+LOOK-M&50\% & 6.89 & 203.47 & 0.70 & 0.62 \\

+HACK&50\% & \textbf{2.06} & \textbf{293.60} & 0.80 & 0.61 \\
\midrule

+Streaming &70\%&4.84 & 228.43& 0.74& 0.62\\
+H2O &70\%& 8.81 & 182.84 & 0.68 & 0.62 \\
+SnapKV&70\% & 7.31& 196.62 & 0.70 & 0.62 \\
+LOOK-M&70\% & 18.88 & 116.70 & 0.58 & 0.63 \\

+HACK &70\%& \textbf{2.78} & \textbf{268.69} & 0.78 & 0.62 \\
\bottomrule
\end{tabular}
\caption{Quantitative comparison of class-conditional image generation on VAR-d30 using the ImageNet dataset.}
\label{tab:varresult}
\end{table}


\textbf{Class-Conditional Image Generation.}
\textit{Quantitative Results:}
HACK exhibits exceptional performance in class-conditional generation tasks, low resolution of which causes sensitivity to compression. In Table\,\ref{tab:varresult}, baseline methods experience great performance deterioration (\emph{e.g.}, LOOK-M’s FID score increases by 18.88 at $70\%$ compression). In contrast, HACK consistently demonstrates superior robustness and outperforms all baselines across different compression ratios, aligning with results from text-to-image generation.

\textit{Qualitative Results:}
The qualitative results in Figure~\ref{varquan} align with our quantitative findings. Images generated by baselines exhibit severe distortions, whereas HACK maintains high visual fidelity, highlighting our effectiveness.

\begin{table}[!t]
\small
\centering

\begin{tabular}{lcccc}
\toprule
Model & Memory & Com. O& Latency &Speedup \\
\midrule
Infinity-2B &  28.96GB & --     & 3.85s  & 1.00× \\

 +HACK ($70\%$)    &14.64GB & 0.16s & 2.61s  & 1.48× \\
 \midrule
Infinity-8B &  60.42GB & --     & 8.14s  & 1.00× \\
 +HACK ($70\%$)    &34.44GB & 0.44s & 5.17s  & 1.57× \\
\midrule
Hart     &   35.47GB & --      & 2.43s  & 1.00× \\
   +HACK ($70\%$)  & 23.82GB  & 0.15s & 1.38s & 1.76× \\
\bottomrule
\end{tabular}
\caption{Efficiency analysis on Infinity and Hart models, evaluated on NVIDIA A100 GPUs with a batch size of 1.}

\label{tab:efficiency}
\end{table}

\subsection{Efficiency Analysis}
%
We evaluate HACK's efficiency by comparing memory consumption and inference latency on Infinity and Hart models under standard attention implementations. Table\,\ref{tab:efficiency} summarizes the results, including additional compression overhead (denoted as Com. O) introduced by HACK. Compared to vanilla VAR models, which suffer from intensive attention computations and cumulative KV cache memory usage, HACK significantly reduces both computational cost and KV cache length, resulting in substantial improvements (\emph{e.g.}, a $1.75\times$ memory reduction and $1.57\times$ speedup on Infinity-8B). Crucially, HACK incurs negligible compression overhead (around $6-8\%$) compared to overall latency.

\begin{figure}[!t]
  \centering
  \includegraphics[width=0.95\linewidth]{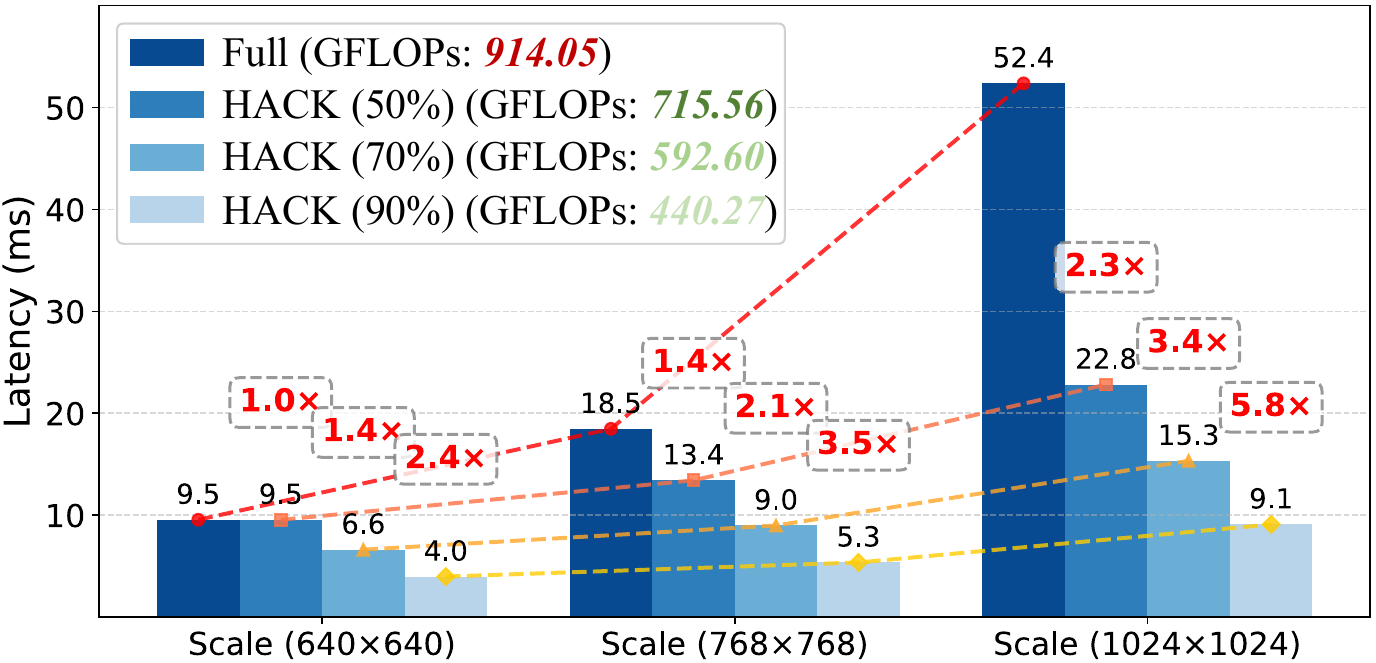}
  \caption{Efficiency Profiling of average latency on attention module for different scales.}
  \label{attn_profile}
\end{figure}

HACK supports efficient scaling to higher resolutions. As shown in Figure\,\ref{attn_profile}, the latency of full attention grows exponentially with increasing resolution, while HACK constrains this growth to be nearly linear, achieving a remarkable $5.8\times$ speedup at 1024$\times$1024 resolution ($\rho=90\%$).
HACK is also compatible with frameworks like FlashAttention~\cite{dao2023flashattention}, yielding further memory savings and speedups. 


\subsection{Ablation Study}

\textbf{Impact of Key Components.}
%
%
Table\,\ref{ablation_key_components} shows that both asymmetric budget allocation and pattern-specific compression are crucial to HACK’s effectiveness. Removing either reduces performance, while combining both yields the best results, highlighting their complementary roles.

\textbf{Ablation Study on Compression Strategies.} 
%
Table\,\ref{ablation_strategy} 
highlights the importance of pattern-specific compression. Removing merging in $f_C$ or deprioritizing key scales in $f_S$ reduces performance, while swapping strategies performs worst, confirming the need for head-type-specific designs. The full method performs best across all metrics.



\textbf{Impact of Query Subset Selection.} Table\,\ref{qselect} evaluates sampling strategies for efficient subset attention in importance estimation. We compare four query selection methods with a fixed subset size ($N_{\text{obs}} = 32$): Random sampling, Initial-$N$ (first $N$ tokens), Recent-$N$ (most recent $N$ tokens), and our uniform sampling. Among them, uniform sampling achieves the closest performance to full attention, offering an effective trade-off between efficiency and generation quality, thereby justifying its use in our framework.


\begin{table}[!t]
\small
\centering

\begin{tabular}{cccccc}
\toprule
Asym.& Pattern &
\multicolumn{2}{c}{Infinity-2B} & 
\multicolumn{2}{c}{VAR-d30} \\
\cmidrule(lr){3-4} \cmidrule(lr){5-6}
Budget&Comp. & IR $\uparrow$ & HPS$\uparrow$ & FID$\downarrow$ & IS $\uparrow$ \\
\midrule
 \ding{55}& \ding{55} &  0.910 & 29.60 & 3.04 & 262.68 \\
\checkmark &\ding{55} &0.914  & 29.58 &2.17 & 288.18 \\
 \ding{55}&\checkmark & 0.923 & 30.08 & 2.19 & 289.24   \\
\checkmark  & \checkmark & \textbf{0.933} & \textbf{30.18} & \textbf{2.06} & \textbf{293.60} \\
\bottomrule
\end{tabular}
\caption{Component ablation of HACK on Infinity-2B ($\rho=70\%$) and VAR-d30 ($\rho=50\%$).}
\label{ablation_key_components}
\end{table}

\begin{table}[!t]
\small
\centering

\begin{tabular}{ccccc}
\toprule
\multirow{2}{*}{Strategy} & 
\multicolumn{2}{c}{Infinity-2B} & 
\multicolumn{2}{c}{VAR-d30} \\
\cmidrule(lr){2-3} \cmidrule(lr){4-5}
 & IR $\uparrow$ & HPS$\uparrow$ & FID$\downarrow$ & IS $\uparrow$ \\
\midrule

$f_C$ w/o merge  &0.931 &30.11& 2.09&291.79\\
$f_S$ w/o init+rec. scales & 0.908& 29.79 &2.17& 287.73\\
$f_C \leftrightarrow f_S$ (Swapped) &0.859 &28.63 &2.69& 271.04\\
Full $f_C$+$f_S$ &\textbf{0.933} &\textbf{30.18} &\textbf{2.06}& \textbf{293.60}\\
\bottomrule
\end{tabular}
\caption{Compression strategy ablation on Infinity-2B ($\rho=70\%$) and VAR-d30 ($\rho=50\%$).}
\label{ablation_strategy}
\end{table}

\begin{table}[!t]
\small
\centering

\begin{tabular}{lccccc}
\toprule
Model & Random & Init & Recent & Uniform & Full \\
\midrule
Infinity-2B &0.927 &0.923 &0.927 &\textbf{0.933} &0.944 \\
Infinity-8B &1.038 &1.028 &1.027 &\textbf{1.043} &1.045 \\
\bottomrule
\end{tabular}
\caption{Query-subset selection strategy ablation on Infinity models evaluated by ImageReward ($\uparrow$) with $\rho=70\%$.}
\label{qselect}
\end{table}

\section{Conclusion}
We propose HACK, a training-free, head-aware KV cache compression framework that makes VAR models efficient. By distinguishing between contextual and structural attention heads, HACK allocates asymmetric cache budgets and applies pattern-specific compression strategies tailored to each head type. HACK enables high compression ratios while preserving generation quality, effectively reducing KV cache size and attention computation, and resulting in substantial memory savings and faster inference. Extensive experiments validate its effectiveness and generalizability.

\section{Acknowledgement}
The paper is supported in part by the National Natural Science Foundation of China (No. U21B2013 , 62325109), and in part by the Shanghai ``The Belt and Road'' Young Scholar Exchange Grant (24510742000).

\bibliography{aaai2026}

@article{achiam2023gpt,
  title={Gpt-4 technical report},
  author={Achiam, Josh and Adler, Steven and Agarwal, Sandhini and Ahmad, Lama and Akkaya, Ilge and Aleman, Florencia Leoni and Almeida, Diogo and Altenschmidt, Janko and Altman, Sam and Anadkat, Shyamal and others},
  journal={arXiv preprint arXiv:2303.08774},
  year={2023}
}

@misc{anthropic2024claude3,
  author = {Anthropic},
  title = {The Claude 3 Model Family: Opus, Sonnet, Haiku},
  year = {2024},
  url = {https://www-cdn.anthropic.com/de8ba9b01c9ab7cbabf5c33b80b7bbc618857627/Model_Card_Claude_3.pdf},
}

@article{dubey2024llama,
  title={The llama 3 herd of models},
  author={Dubey, Abhimanyu and Jauhri, Abhinav and Pandey, Abhinav and Kadian, Abhishek and Al-Dahle, Ahmad and Letman, Aiesha and Mathur, Akhil and Schelten, Alan and Yang, Amy and Fan, Angela and others},
  journal={arXiv preprint arXiv:2407.21783},
  year={2024}
}

@article{liu2024deepseek,
  title={Deepseek-v3 technical report},
  author={Liu, Aixin and Feng, Bei and Xue, Bing and Wang, Bingxuan and Wu, Bochao and Lu, Chengda and Zhao, Chenggang and Deng, Chengqi and Zhang, Chenyu and Ruan, Chong and others},
  journal={arXiv preprint arXiv:2412.19437},
  year={2024}
}

@inproceedings{ho2020denoising,
  title={Denoising diffusion probabilistic models},
  author={Ho, Jonathan and Jain, Ajay and Abbeel, Pieter},
  booktitle={Advances in neural information processing systems},
  pages={6840--6851},
  year={2020}
}

@article{podell2023sdxl,
  title={Sdxl: Improving latent diffusion models for high-resolution image synthesis},
  author={Podell, Dustin and English, Zion and Lacey, Kyle and Blattmann, Andreas and Dockhorn, Tim and M{\"u}ller, Jonas and Penna, Joe and Rombach, Robin},
  journal={arXiv preprint arXiv:2307.01952},
  year={2023}
}

@inproceedings{peebles2023scalable,
  title={Scalable diffusion models with transformers},
  author={Peebles, William and Xie, Saining},
  booktitle={Proceedings of the IEEE/CVF international conference on computer vision},
  pages={4195--4205},
  year={2023}
}

@inproceedings{esser2024scaling,
  title={Scaling rectified flow transformers for high-resolution image synthesis},
  author={Esser, Patrick and Kulal, Sumith and Blattmann, Andreas and Entezari, Rahim and M{\"u}ller, Jonas and Saini, Harry and Levi, Yam and Lorenz, Dominik and Sauer, Axel and Boesel, Frederic and others},
  booktitle={Forty-first international conference on machine learning},
  year={2024}
}

@inproceedings{chen2024pixart,
  title={Pixart-$\sigma$: Weak-to-strong training of diffusion transformer for 4k text-to-image generation},
  author={Chen, Junsong and Ge, Chongjian and Xie, Enze and Wu, Yue and Yao, Lewei and Ren, Xiaozhe and Wang, Zhongdao and Luo, Ping and Lu, Huchuan and Li, Zhenguo},
  booktitle={European Conference on Computer Vision},
  pages={74--91},
  year={2024}
}

@article{he2024mars,
  title={Mars: Mixture of auto-regressive models for fine-grained text-to-image synthesis},
  author={He, Wanggui and Fu, Siming and Liu, Mushui and Wang, Xierui and Xiao, Wenyi and Shu, Fangxun and Wang, Yi and Zhang, Lei and Yu, Zhelun and Li, Haoyuan and others},
  journal={arXiv preprint arXiv:2407.07614},
  year={2024}
}

@article{sun2024autoregressive,
  title={Autoregressive model beats diffusion: Llama for scalable image generation},
  author={Sun, Peize and Jiang, Yi and Chen, Shoufa and Zhang, Shilong and Peng, Bingyue and Luo, Ping and Yuan, Zehuan},
  journal={arXiv preprint arXiv:2406.06525},
  year={2024}
}

@article{li2024autoregressive,
  title={Autoregressive image generation without vector quantization},
  author={Li, Tianhong and Tian, Yonglong and Li, He and Deng, Mingyang and He, Kaiming},
  journal={Advances in Neural Information Processing Systems},
  volume={37},
  pages={56424--56445},
  year={2024}
}

@article{tian2024visual,
  title={Visual autoregressive modeling: Scalable image generation via next-scale prediction},
  author={Tian, Keyu and Jiang, Yi and Yuan, Zehuan and Peng, Bingyue and Wang, Liwei},
  journal={Advances in neural information processing systems},
  volume={37},
  pages={84839--84865},
  year={2024}
}

@article{han2024infinity,
  title={Infinity: Scaling bitwise autoregressive modeling for high-resolution image synthesis},
  author={Han, Jian and Liu, Jinlai and Jiang, Yi and Yan, Bin and Zhang, Yuqi and Yuan, Zehuan and Peng, Bingyue and Liu, Xiaobing},
  journal={arXiv preprint arXiv:2412.04431},
  year={2024}
}

@article{tang2024hart,
  title={Hart: Efficient visual generation with hybrid autoregressive transformer},
  author={Tang, Haotian and Wu, Yecheng and Yang, Shang and Xie, Enze and Chen, Junsong and Chen, Junyu and Zhang, Zhuoyang and Cai, Han and Lu, Yao and Han, Song},
  journal={arXiv preprint arXiv:2410.10812},
  year={2024}
}

@article{van2017neural,
  title={Neural discrete representation learning},
  author={Van Den Oord, Aaron and Vinyals, Oriol and others},
  journal={Advances in neural information processing systems},
  volume={30},
  year={2017}
}

@inproceedings{esser2021taming,
  title={Taming transformers for high-resolution image synthesis},
  author={Esser, Patrick and Rombach, Robin and Ommer, Bjorn},
  booktitle={Proceedings of the IEEE/CVF conference on computer vision and pattern recognition},
  pages={12873--12883},
  year={2021}
}

@article{chen2024toward,
  title={Toward Guidance-Free AR Visual Generation via Condition Contrastive Alignment},
  author={Chen, Huayu and Su, Hang and Sun, Peize and Zhu, Jun},
  journal={arXiv preprint arXiv:2410.09347},
  year={2024}
}

@article{ren2024m,
  title={M-VAR: Decoupled Scale-wise Autoregressive Modeling for High-Quality Image Generation},
  author={Ren, Sucheng and Yu, Yaodong and Ruiz, Nataniel and Wang, Feng and Yuille, Alan and Xie, Cihang},
  journal={arXiv preprint arXiv:2411.10433},
  year={2024}
}

@inproceedings{voronov2024switti,
  title={Switti: Designing Scale-Wise Transformers for Text-to-Image Synthesis},
  author={Voronov, Anton and Kuznedelev, Denis and Khoroshikh, Mikhail and Khrulkov, Valentin and Baranchuk, Dmitry},
  journal={arXiv preprint arXiv:2412.01819},
  year={2024}
}

@inproceedings{chen2024sar3d,
    title={SAR3D: Autoregressive 3D Object Generation and Understanding via Multi-scale 3D VQVAE},
    author={Chen, Yongwei and Lan, Yushi and Zhou, Shangchen and Wang, Tengfei and Pan, Xingang},
    booktitle={CVPR},
    year={2025}
}

@article{gao2025mars,
  title={MARS: Mesh AutoRegressive Model for 3D Shape Detailization},
  author={Gao, Jingnan and Liu, Weizhe and Sun, Weixuan and Wang, Senbo and Song, Xibin and Shang, Taizhang and Chen, Shenzhou and Li, Hongdong and Yang, Xiaokang and Yan, Yichao and others},
  journal={arXiv preprint arXiv:2502.11390},
  year={2025}
}

@article{zhuang2025vargpt,
  title={Vargpt-v1. 1: Improve visual autoregressive large unified model via iterative instruction tuning and reinforcement learning},
  author={Zhuang, Xianwei and Xie, Yuxin and Deng, Yufan and Yang, Dongchao and Liang, Liming and Ru, Jinghan and Yin, Yuguo and Zou, Yuexian},
  journal={arXiv preprint arXiv:2504.02949},
  year={2025}
}

@article{xiao2023efficient,
  title={Efficient streaming language models with attention sinks},
  author={Xiao, Guangxuan and Tian, Yuandong and Chen, Beidi and Han, Song and Lewis, Mike},
  journal={arXiv preprint arXiv:2309.17453},
  year={2023}
}

@article{zhang2024h2o,
  title={H2o: Heavy-hitter oracle for efficient generative inference of large language models},
  author={Zhang, Zhenyu and Sheng, Ying and Zhou, Tianyi and Chen, Tianlong and Zheng, Lianmin and Cai, Ruisi and Song, Zhao and Tian, Yuandong and R{\'e}, Christopher and Barrett, Clark and others},
  journal={Advances in Neural Information Processing Systems},
  volume={36},
  year={2024}
}

@article{liu2024scissorhands,
  title={Scissorhands: Exploiting the persistence of importance hypothesis for llm kv cache compression at test time},
  author={Liu, Zichang and Desai, Aditya and Liao, Fangshuo and Wang, Weitao and Xie, Victor and Xu, Zhaozhuo and Kyrillidis, Anastasios and Shrivastava, Anshumali},
  journal={Advances in Neural Information Processing Systems},
  volume={36},
  year={2024}
}

@article{oren2024transformers,
  title={Transformers are multi-state rnns},
  author={Oren, Matanel and Hassid, Michael and Adi, Yossi and Schwartz, Roy},
  journal={arXiv preprint arXiv:2401.06104},
  year={2024}
}

@article{ren2024efficacy,
  title={On the efficacy of eviction policy for key-value constrained generative language model inference},
  author={Ren, Siyu and Zhu, Kenny Q},
  journal={arXiv preprint arXiv:2402.06262},
  year={2024}
}

@article{li2024snapkv,
  title={Snapkv: Llm knows what you are looking for before generation},
  author={Li, Yuhong and Huang, Yingbing and Yang, Bowen and Venkitesh, Bharat and Locatelli, Acyr and Ye, Hanchen and Cai, Tianle and Lewis, Patrick and Chen, Deming},
  journal={arXiv preprint arXiv:2404.14469},
  year={2024}
}

@article{ge2023model,
  title={Model tells you what to discard: Adaptive kv cache compression for llms},
  author={Ge, Suyu and Zhang, Yunan and Liu, Liyuan and Zhang, Minjia and Han, Jiawei and Gao, Jianfeng},
  journal={arXiv preprint arXiv:2310.01801},
  year={2023}
}

@article{yang2024pyramidinfer,
  title={PyramidInfer: Pyramid KV Cache Compression for High-throughput LLM Inference},
  author={Yang, Dongjie and Han, XiaoDong and Gao, Yan and Hu, Yao and Zhang, Shilin and Zhao, Hai},
  journal={arXiv preprint arXiv:2405.12532},
  year={2024}
}

@article{wan2024look,
  title={Look-m: Look-once optimization in kv cache for efficient multimodal long-context inference},
  author={Wan, Zhongwei and Wu, Ziang and Liu, Che and Huang, Jinfa and Zhu, Zhihong and Jin, Peng and Wang, Longyue and Yuan, Li},
  journal={arXiv preprint arXiv:2406.18139},
  year={2024}
}

@article{feng2024ada,
  title={Ada-kv: Optimizing kv cache eviction by adaptive budget allocation for efficient llm inference},
  author={Feng, Yuan and Lv, Junlin and Cao, Yukun and Xie, Xike and Zhou, S Kevin},
  journal={arXiv preprint arXiv:2407.11550},
  year={2024}
}

@article{fu2024not,
  title={Not all heads matter: A head-level KV cache compression method with integrated retrieval and reasoning},
  author={Fu, Yu and Cai, Zefan and Asi, Abedelkadir and Xiong, Wayne and Dong, Yue and Xiao, Wen},
  journal={arXiv preprint arXiv:2410.19258},
  year={2024}
}

@article{qin2025cake,
  title={Cake: Cascading and adaptive kv cache eviction with layer preferences},
  author={Qin, Ziran and Cao, Yuchen and Lin, Mingbao and Hu, Wen and Fan, Shixuan and Cheng, Ke and Lin, Weiyao and Li, Jianguo},
  journal={arXiv preprint arXiv:2503.12491},
  year={2025}
}

@article{wan2024d2o,
  title={D2O: Dynamic Discriminative Operations for Efficient Generative Inference of Large Language Models},
  author={Wan, Zhongwei and Wu, Xinjian and Zhang, Yu and Xin, Yi and Tao, Chaofan and Zhu, Zhihong and Wang, Xin and Luo, Siqi and Xiong, Jing and Zhang, Mi},
  journal={arXiv preprint arXiv:2406.13035},
  year={2024}
}

@article{liu2024kivi,
  title={Kivi: A tuning-free asymmetric 2bit quantization for kv cache},
  author={Liu, Zirui and Yuan, Jiayi and Jin, Hongye and Zhong, Shaochen and Xu, Zhaozhuo and Braverman, Vladimir and Chen, Beidi and Hu, Xia},
  journal={arXiv preprint arXiv:2402.02750},
  year={2024}
}

@article{yue2024wkvquant,
  title={Wkvquant: Quantizing weight and key/value cache for large language models gains more},
  author={Yue, Yuxuan and Yuan, Zhihang and Duanmu, Haojie and Zhou, Sifan and Wu, Jianlong and Nie, Liqiang},
  journal={arXiv preprint arXiv:2402.12065},
  year={2024}
}

@article{kang2024gear,
  title={Gear: An efficient kv cache compression recipefor near-lossless generative inference of llm},
  author={Kang, Hao and Zhang, Qingru and Kundu, Souvik and Jeong, Geonhwa and Liu, Zaoxing and Krishna, Tushar and Zhao, Tuo},
  journal={arXiv preprint arXiv:2403.05527},
  year={2024}
}

@article{he2024zipcache,
  title={ZipCache: Accurate and Efficient KV Cache Quantization with Salient Token Identification},
  author={He, Yefei and Zhang, Luoming and Wu, Weijia and Liu, Jing and Zhou, Hong and Zhuang, Bohan},
  journal={arXiv preprint arXiv:2405.14256},
  year={2024}
}

@article{wan2025meda,
  title={Meda: Dynamic kv cache allocation for efficient multimodal long-context inference},
  author={Wan, Zhongwei and Shen, Hui and Wang, Xin and Liu, Che and Mai, Zheda and Zhang, Mi},
  journal={arXiv preprint arXiv:2502.17599},
  year={2025}
}

@inproceedings{zhang2024cam,
  title={Cam: Cache merging for memory-efficient llms inference},
  author={Zhang, Yuxin and Du, Yuxuan and Luo, Gen and Zhong, Yunshan and Zhang, Zhenyu and Liu, Shiwei and Ji, Rongrong},
  booktitle={Forty-first International Conference on Machine Learning},
  year={2024}
}

@article{liu2024minicache,
  title={MiniCache: KV Cache Compression in Depth Dimension for Large Language Models},
  author={Liu, Akide and Liu, Jing and Pan, Zizheng and He, Yefei and Haffari, Gholamreza and Zhuang, Bohan},
  journal={arXiv preprint arXiv:2405.14366},
  year={2024}
}

@article{xie2024litevar,
  title={LiteVAR: Compressing Visual Autoregressive Modelling with Efficient Attention and Quantization},
  author={Xie, Rui and Zhao, Tianchen and Yuan, Zhihang and Wan, Rui and Gao, Wenxi and Zhu, Zhenhua and Ning, Xuefei and Wang, Yu},
  journal={arXiv preprint arXiv:2411.17178},
  year={2024}
}

@article{dao2023flashattention,
  title={Flashattention-2: Faster attention with better parallelism and work partitioning},
  author={Dao, Tri},
  journal={arXiv preprint arXiv:2307.08691},
  year={2023}
}

@article{ghosh2023geneval,
  title={Geneval: An object-focused framework for evaluating text-to-image alignment},
  author={Ghosh, Dhruba and Hajishirzi, Hannaneh and Schmidt, Ludwig},
  journal={Advances in Neural Information Processing Systems},
  pages={52132--52152},
  year={2023}
}

@article{xu2023imagereward,
  title={Imagereward: Learning and evaluating human preferences for text-to-image generation},
  author={Xu, Jiazheng and Liu, Xiao and Wu, Yuchen and Tong, Yuxuan and Li, Qinkai and Ding, Ming and Tang, Jie and Dong, Yuxiao},
  journal={Advances in Neural Information Processing Systems},
  pages={15903--15935},
  year={2023}
}

@article{li2024playground,
  title={Playground v2. 5: Three insights towards enhancing aesthetic quality in text-to-image generation},
  author={Li, Daiqing and Kamko, Aleks and Akhgari, Ehsan and Sabet, Ali and Xu, Linmiao and Doshi, Suhail},
  journal={arXiv preprint arXiv:2402.17245},
  year={2024}
}

@article{wu2023human,
  title={Human preference score v2: A solid benchmark for evaluating human preferences of text-to-image synthesis},
  author={Wu, Xiaoshi and Hao, Yiming and Sun, Keqiang and Chen, Yixiong and Zhu, Feng and Zhao, Rui and Li, Hongsheng},
  journal={arXiv preprint arXiv:2306.09341},
  year={2023}
}

@inproceedings{deng2009imagenet,
  title={Imagenet: A large-scale hierarchical image database},
  author={Deng, Jia and Dong, Wei and Socher, Richard and Li, Li-Jia and Li, Kai and Fei-Fei, Li},
  booktitle={IEEE conference on computer vision and pattern recognition},
  pages={248--255},
  year={2009}
}

@article{guo2025fastvar,
  title={Fastvar: Linear visual autoregressive modeling via cached token pruning},
  author={Guo, Hang and Li, Yawei and Zhang, Taolin and Wang, Jiangshan and Dai, Tao and Xia, Shu-Tao and Benini, Luca},
  journal={arXiv preprint arXiv:2503.23367},
  year={2025}
}

@article{qin2025autoregressive,
  title={Autoregressive Image Generation Needs Only a Few Lines of Cached Tokens},
  author={Qin, Ziran and Lv, Youru and Lin, Mingbao and Zhang, Zeren and Gan, Chanfan and Chen, Tieyuan and Lin, Weiyao},
  journal={arXiv preprint arXiv:2512.04857},
  year={2025}
}

@article{wu2024janus,
  title={Janus: Decoupling visual encoding for unified multimodal understanding and generation},
  author={Wu, Chengyue and Chen, Xiaokang and Wu, Zhiyu and Ma, Yiyang and Liu, Xingchao and Pan, Zizheng and Liu, Wen and Xie, Zhenda and Yu, Xingkai and Ruan, Chong and others},
  journal={arXiv preprint arXiv:2410.13848},
  year={2024}
}

@misc{liu2024lumina-mgpt,
      title={Lumina-mGPT: Illuminate Flexible Photorealistic Text-to-Image Generation with Multimodal Generative Pretraining},
      author={Dongyang Liu and Shitian Zhao and Le Zhuo and Weifeng Lin and Yu Qiao and Hongsheng Li and Peng Gao},
      year={2024},
      eprint={2408.02657},
      archivePrefix={arXiv},
      primaryClass={cs.CV},
      url={https://arxiv.org/abs/2408.02657},
}
\newpage
\appendix

\section{Implementation Details}
\subsection{Experimental Setting}
For all evaluated models, including Infinity-2B/8B~\cite {han2024infinity}, Hart~\cite{tang2024hart}, and VAR-d30~\cite{tian2024visual}, we strictly follow their official evaluation protocols, including hyperparameter configurations and experimental environments, to ensure fair and accurate assessment.
\begin{itemize}
    \item Text-to-Image Generation: For models focused on this task (Infinity-2B/8B, and Hart), we generate images at a resolution of 1024$\times$1024 for evaluation, in line with their standard settings.
    \item Class-Conditional Generation: For the VAR-d30 model, we follow its benchmark for class-conditional image generation, producing images at 256$\times$256 resolution for evaluation.
\end{itemize}
To ensure reproducibility and eliminate the effects of randomness, we fix the random seed across all experiments and baseline comparisons.

\subsection{HACK Implementation Details}
\textbf{General Settings.}
\textit{Offline head classification Setting}. 
We perform a one-time, offline head classification for each model using only 50 evaluation samples, leveraging the stability of attention patterns across inputs and scales. (More detailed ablation studies on head classification are provided in Section\,\ref {ssc}.) Specifically, we use prompts from the ImageReward benchmark~\cite{xu2023imagereward} for Infinity-2B/8B and Hart; class labels from ImageNet~\cite{deng2009imagenet} for VAR-d30. \textbf{The entire head classification process completes within minutes.} To enable efficient inference, we statically reorder heads within each layer to group contextual and structural heads before deployment.

\textit{Compression Setting.}
To reduce computational overhead, we apply subset attention with a fixed subset size of $N_{\text{obs}} = 32$, uniformly sampled from the full queries to estimate token importance across all models. For the contextual head compression strategy $f_C$, we apply an additional merging operation only at the final generation step to preserve more semantic information. For the structural head compression strategy $f_S$, we consistently retain tokens from the first two scales and the most recent scale to preserve spatial continuity. Evicted KV pairs will not be recomputed or reused in later steps.

\textbf{Model-Specific Configurations.}
We moderately adjust the contextual head ratio $\alpha$ and the budget allocation between contextual ($B_C$) and structural ($B_S$) heads to better align with each model's head variance distribution and compression sensitivity. These configurations adhere to our asymmetric budget allocation principle, governed by the equation $B=\alpha B_C + (1-\alpha) B_S$, where $B$ is the target average budget.
For high-resolution generation models such as Infinity-2B, Infinity-8B,  and Hart, we set $\alpha$ to 0.1, 0.3, and 0.3, respectively. We allocate the budgets with a fixed ratio of $B_C : B_S = 1 : 2$ to ensure that structural heads receive a larger share of the cache budget.
For the low-resolution generation model VAR-d30, we set $\alpha = 0.35$, and directly specify the contextual head budget as $B_C = 0.1 \cdot T_K$ and $B_C = 0.05 \cdot T_K$, respectively, thereby reserving more capacity for compression-sensitive structural heads, where $T_K$ denotes the total token number.

\textbf{Importantly, HACK does not rely on extensive hyperparameter tuning.} As shown in the sensitivity analysis (see Section\,\ref{sensitystudy}), it consistently delivers strong performance across a wide range of $\alpha$ values and budget settings.

\begin{figure}[h]
  \centering
  \subfigure[Sensitivity study on contextual head ratio $\alpha$]{
    \includegraphics[width=0.95\linewidth]{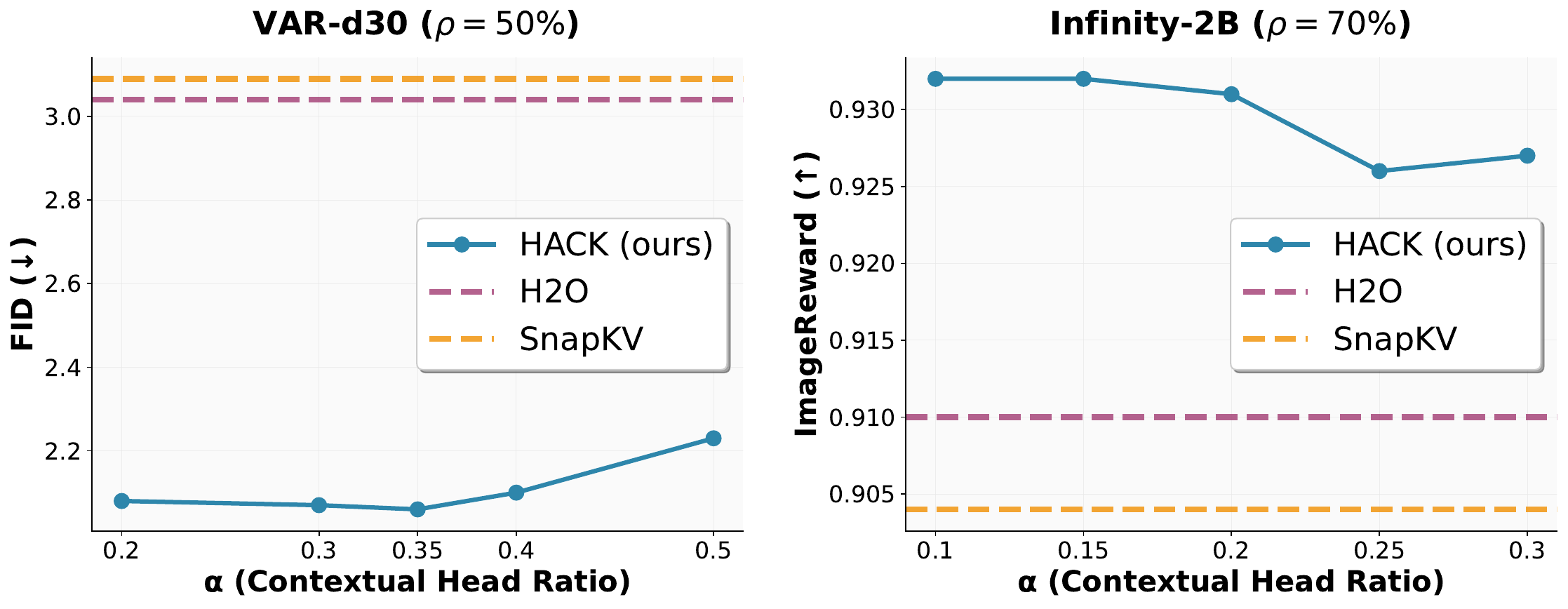}
    \label{fig:Head_classification_infinity}
  }
  \subfigure[Sensitivity study on contextual head budget ratio $B_C/T_K$]{
    \includegraphics[width=0.95\linewidth]{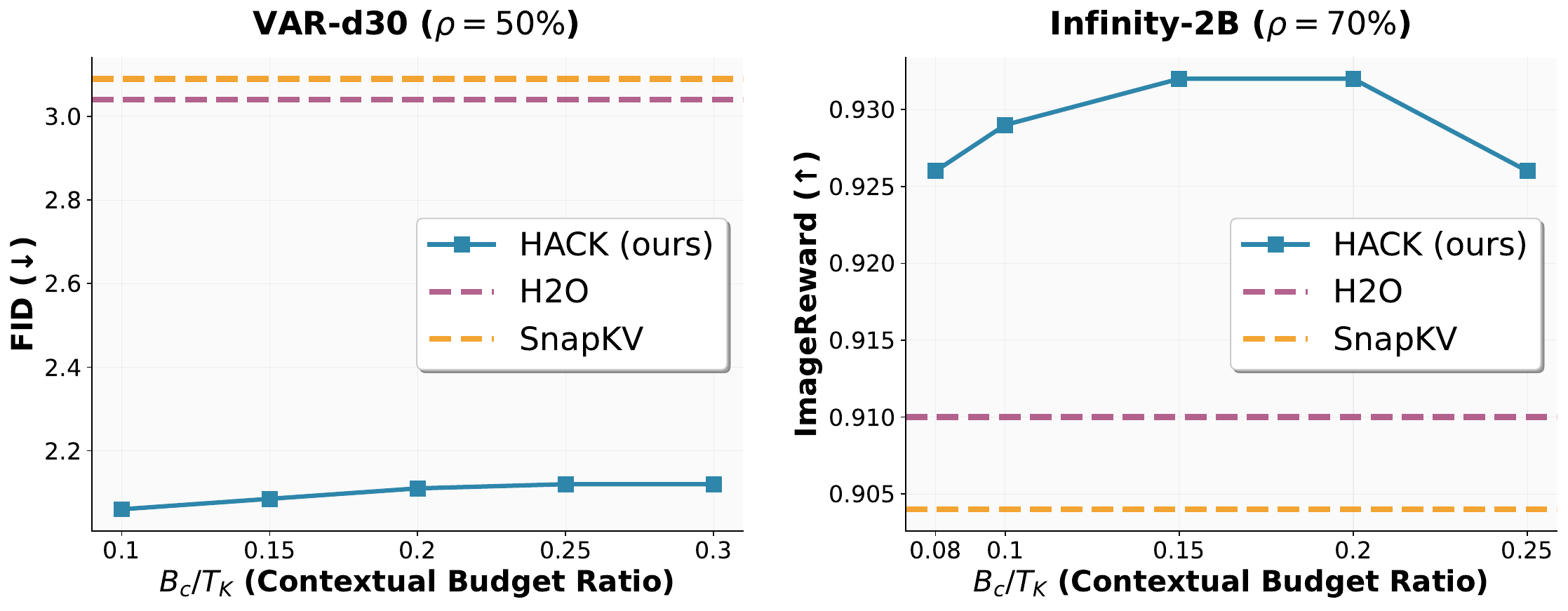}
    \label{fig:Head_classification_var}
  }
  \caption{Sensitivity study on Hyperparameters. We compare HACK with different hyperparameter settings with two strong baselines on VAR-d30 and Infinity-2B, respectively.}
  \label{fig:hyper_sensi}
\end{figure}

\section{Additional Ablation Study}
\subsection{Sensitivity Study on Hyperparameters}
\label{sensitystudy}
We explore the influence of parameter configuration on HACK by evaluating its performance under varying settings of its two hyperparameters: the contextual head ratio $\alpha$ and the budget allocation for contextual and structural heads that can be parameterized by the contextual budget ratio $B_C/T_K$. 
We conduct two sensitivity studies: (a) fixing the contextual budget ratio $B_C/T_K$ while varying the contextual head ratio $\alpha$, and (b) fixing $\alpha$ while varying $B_C/T_K$. We perform these experiments on two representative models, VAR-d30 and Infinity-2B.
As shown in Figure\,\ref{fig:hyper_sensi}, HACK consistently outperforms both strong baselines (H2O and SnapKV) across a wide range of settings, demonstrating its robustness to hyperparameter choices.
While the performance of HACK remains consistently strong, we observe that a more balanced hyperparameter setting (e.g., $\alpha=0.35$, $B_C/T_K=0.1$ in VAR-d30) can lead to further improvements. This suggests that HACK does not rely on extensive hyperparameter tuning, but can benefit from reasonable configurations, which allows it to be flexibly adapted to different deployment scenarios.

\subsection{Ablation Study on Head Classification}
\textbf{Effectiveness of Attention Variance.} 
To verify whether attention variance provides a meaningful signal for head classification, we compare our variance-based strategy against three alternative heuristics: (1) \textit{Order}, which classifies heads sequentially within each layer; (2) \textit{Uniform}, which evenly splits heads in every layer; and (3) \textit{Random}, which assigns contextual heads randomly while preserving the global $\alpha$ ratio. As shown in Table~\ref{headclassify}, our variance-based classification achieves significantly better performance, confirming its superiority. This demonstrates that attention variance is an effective indicator of functional head roles, aligning well with the observed contextual and structural attention patterns.

\begin{table}[!t]
\centering

\begin{tabular}{lcccccc}
\toprule
Method& FID↓ & IS$\uparrow$ & Precision$\uparrow$ & Recall$\uparrow$ \\

 \midrule
Order &2.57  & 273.29 & 0.78 & 0.62 \\
Uniform &2.63  & 272.79 & 0.78 & 0.61 \\
Random & 2.70 & 270.47 & 0.78 & 0.61 \\
Variance & \textbf{2.06}  &  \textbf{293.60} & 0.80 & 0.61 \\

\bottomrule
\end{tabular}
\caption{Comparison with different head classification strategies on VAR-d30 with $\alpha=0.35$ and compression ratio $\rho=50\%$.}
\label{headclassify}
\end{table}

\begin{figure}[!t]
  \centering
  \includegraphics[width=\linewidth]{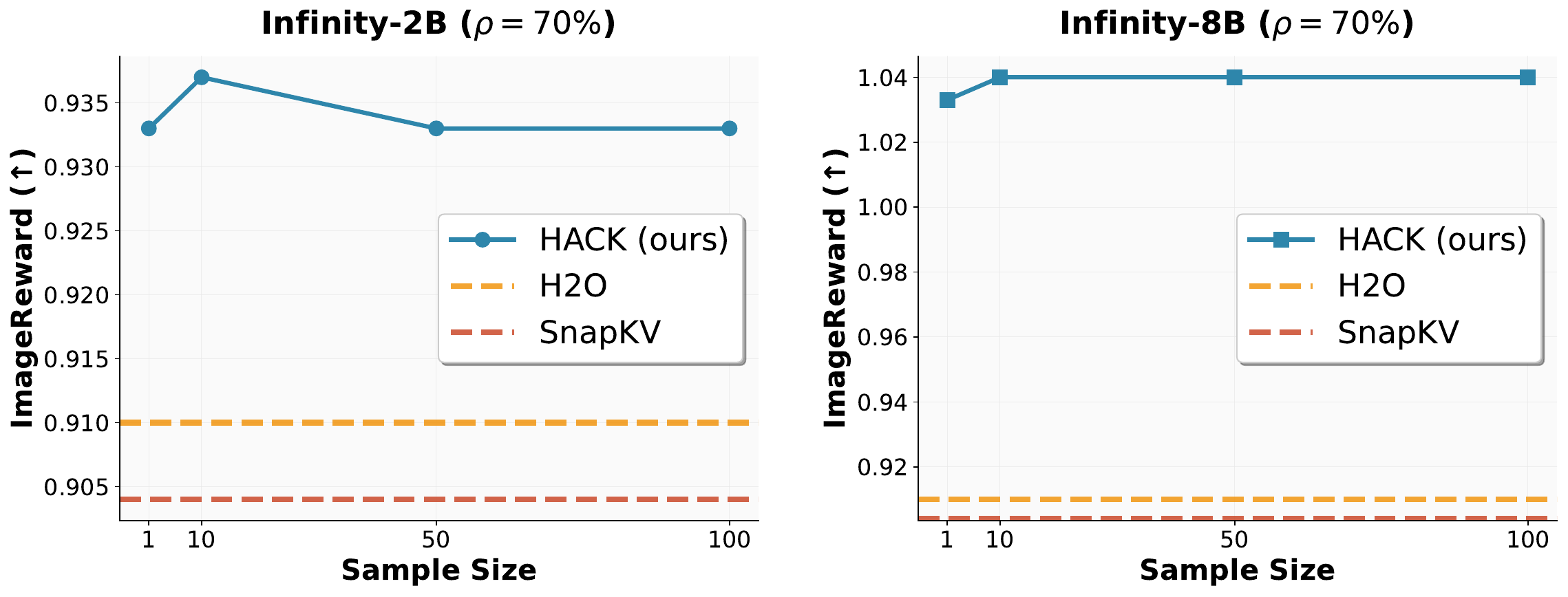}   
  \caption{Performance consistency of HACK on Infinity-2B and -8B. The models are configured using different sample sizes (from 1 to 100) for the offline head classification, yet they all achieve a consistently high ImageReward score in the final evaluation.}
    \label{fig:sampleab}
\end{figure}

\textbf{Impact of Sample Size on Head Classification}\label{ssc}
We investigate the impact of the sample size used for offline head classification. As visualized in Figure~\ref{fig:Head_classification}, the classification results for both Infinity-8B and VAR-d30 remain remarkably stable across varying numbers of prompts (from 1 to 100) or class labels (from 1 to 500). This stability directly translates into consistently strong model performance, as evidenced by the flat ImageReward curves in Figure~\ref{fig:sampleab}.
These results confirm our core observation: the functional roles of attention heads are largely invariant to the specific samples used for analysis, suggesting a strong inductive prior in head behavior. This enables us to perform accurate classification using a small number of samples. We therefore set the default sample size to 50, which strikes a practical balance between reliability and computational efficiency.

\begin{figure}[!t]
  \centering
  \includegraphics[width=\linewidth]{Head_classification.pdf}   \caption{Head classification results of Infinity-8B and VAR models with varying sample sizes. Yellow indicates Contextual Heads (low variance); purple indicates Structural Heads (high variance).
  \label{fig:Head_classification}}
\end{figure}

\subsection{Impact of Query Sampling Size on Subset Attention}
Table~\ref{tab:ablation_query_subset} presents the impact of query sampling size $N_\text{obs}$ on the effectiveness of subset attention. As $N_\text{obs}$ increases, the compressed model demonstrates progressively improved performance, approaching that of full attention. This is because a larger query subset allows for more accurate estimation of token importance. Remarkably, when $N_\text{obs}=32$, the model achieves performance nearly equivalent to the full attention baseline, striking a favorable balance between accuracy and efficiency. We adopt $N_\text{obs}=32$  across all models.

\begin{table}[!t]
\small
\centering

\setlength{\tabcolsep}{4pt}
\begin{tabular}{lccccr}
\toprule
$\rho$ & $N_{\text{obs}}$ = 4 & $N_{\text{obs}}$ = 8 & $N_{\text{obs}}$ = 16 & $N_{\text{obs}}$ = 32 & Full \\
\midrule
70\% &0.923 &0.928 &0.929 &0.933 &0.931 \\
90\% &0.800 &0.856 &0.843 &0.862 &0.862 \\
\bottomrule
\end{tabular}
\caption{Impact of query sampling size $N_{\text{obs}}$ on subset attention. The experiments are conducted on Infinity-2B using ImageReward ($\uparrow$).}
\label{tab:ablation_query_subset}
\end{table}

\section{More Visual Results}
\subsection{Additional Qualitative Results}
\begin{figure*}[t]
  \centering
  \includegraphics[width=\linewidth]
  {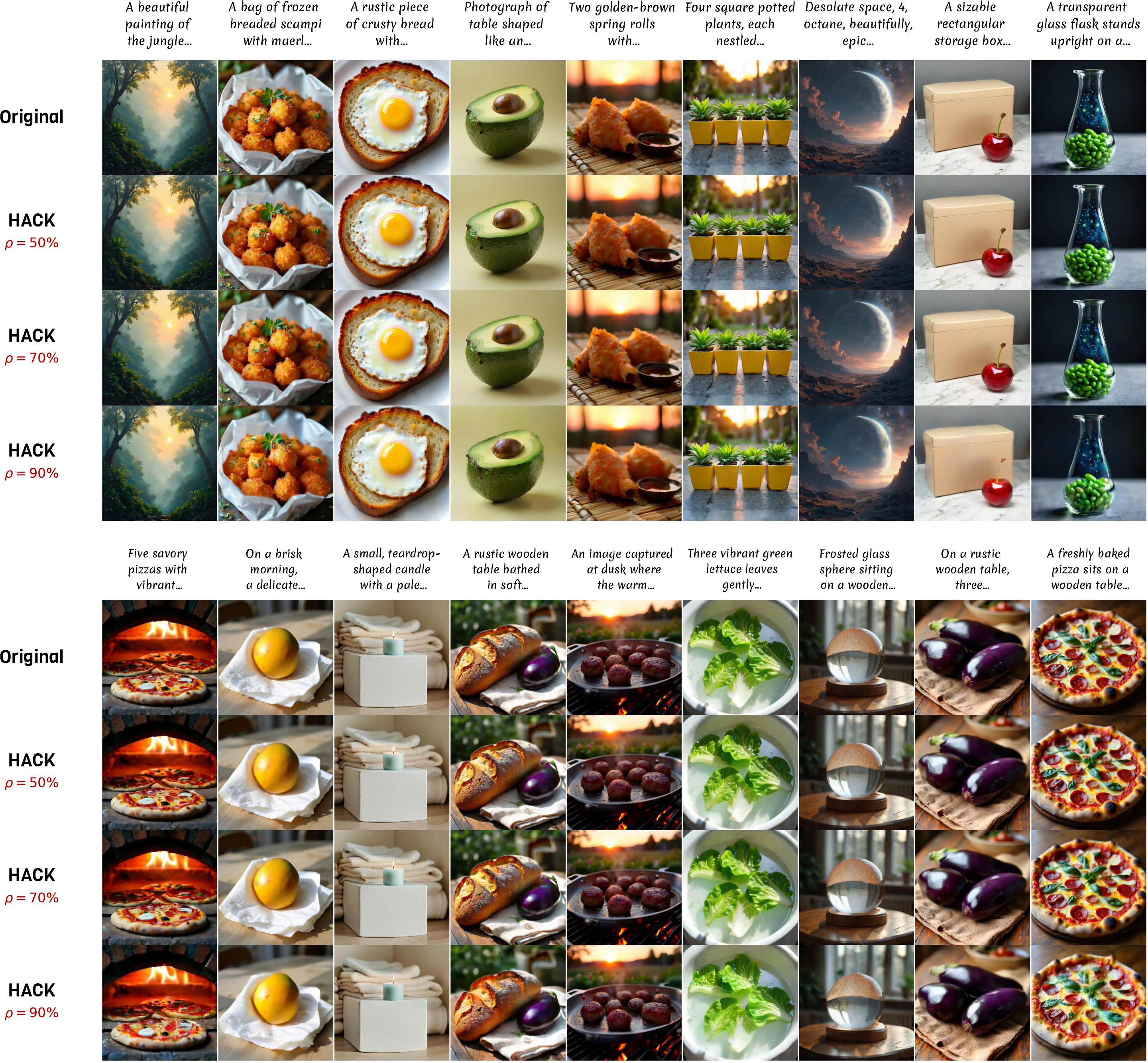}
  \caption{Qualitative comparison of images generated by Infinity-2B under different compression ratios using HACK.}
  \label{fig:2b_more_sup}
\end{figure*}

\begin{figure*}[t]
  \centering
  \includegraphics[width=\linewidth]
  {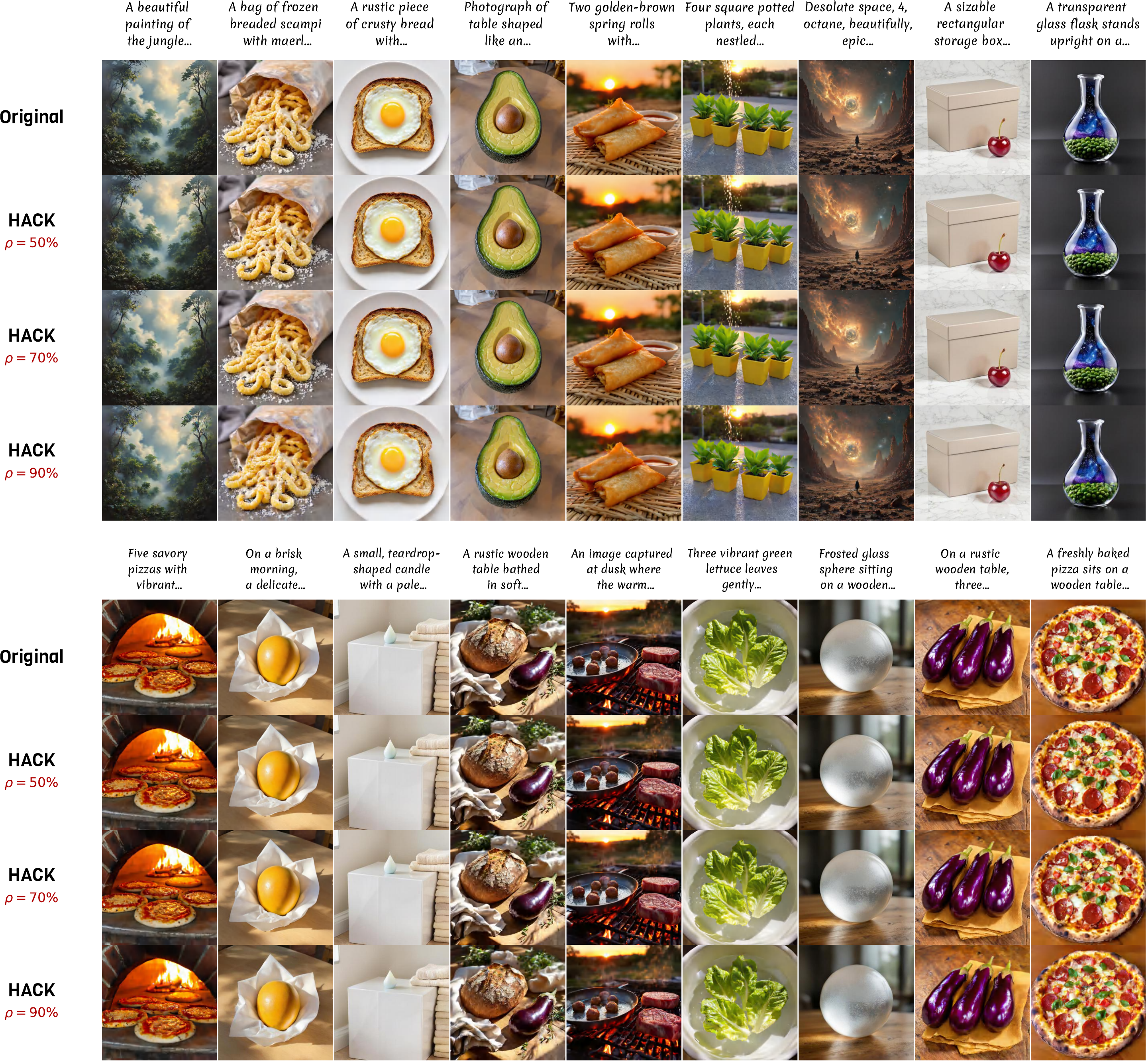}
  \caption{Qualitative comparison of images generated by Infinity-8B under different compression ratios using HACK. }
  \label{fig:8b_more_sup}
\end{figure*}

\begin{figure*}[t]
  \centering
  \includegraphics[width=\linewidth]
  {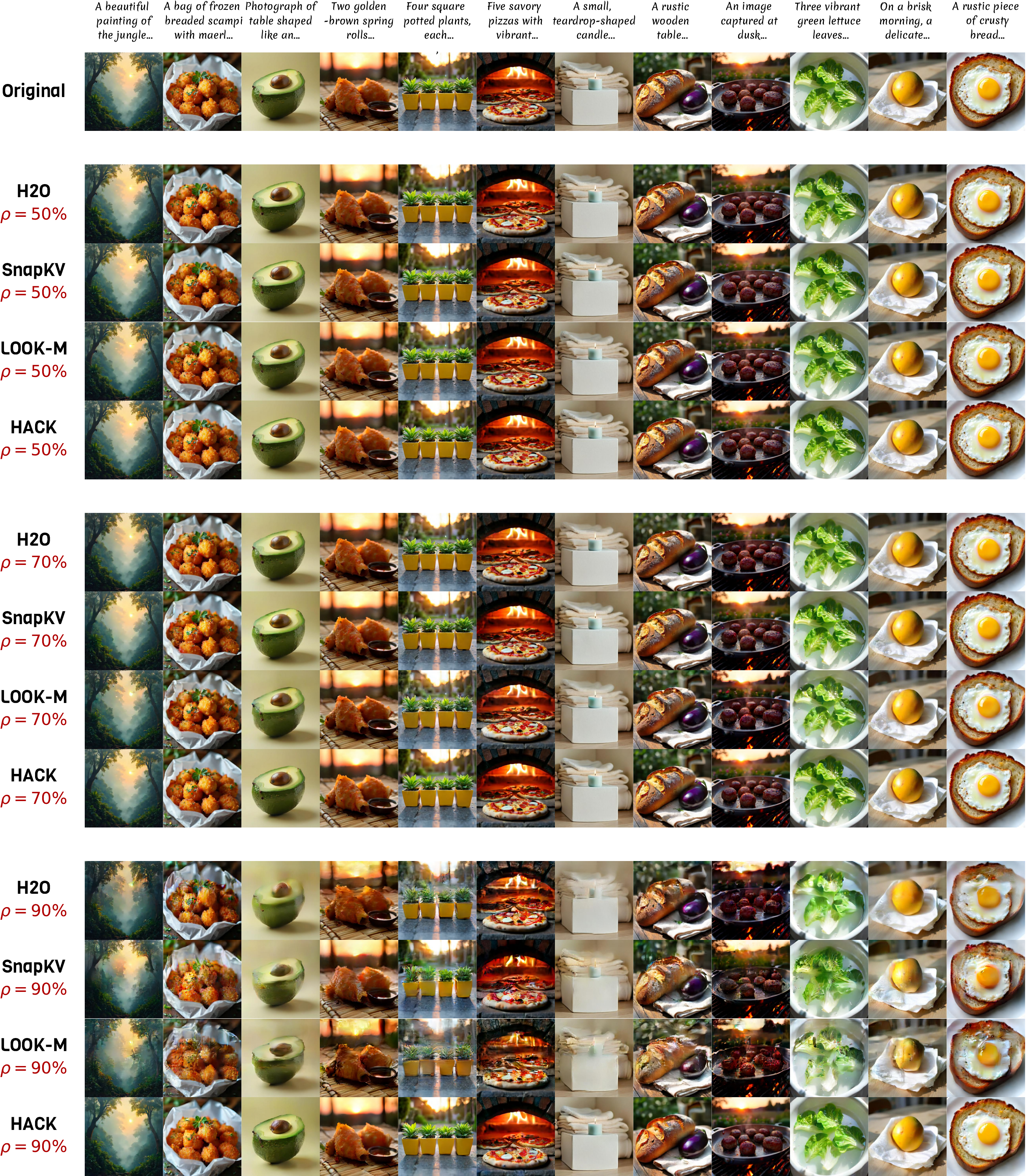}
  \caption{Comparison between HACK and baseline methods on Infinity-2B under varying compression ratios.}
  \label{fig:2b_sup}
\end{figure*}

\begin{figure*}[t]
  \centering
  \includegraphics[width=\linewidth]
  {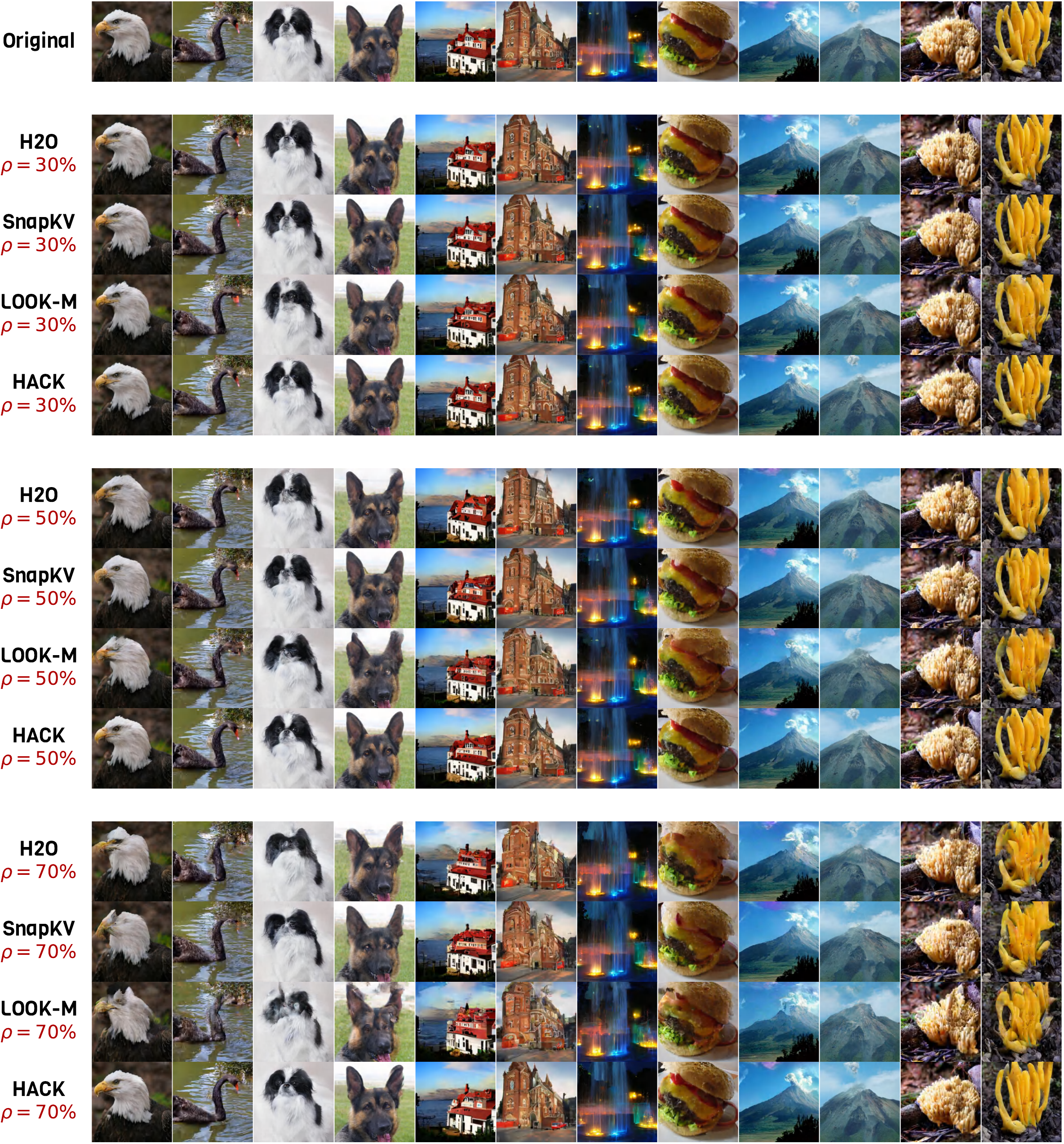}
  \caption{Comparison between HACK and baseline methods on VAR-d30 under varying compression ratios.}
  \label{fig:var_sup}
\end{figure*}

We provide additional qualitative results on Infinity-2B/8B and VAR-d30 models. 
Figures~\ref{fig:2b_more_sup} and~\ref{fig:8b_more_sup} show generation results at varying compression ratios $\rho$ on Infinity-2B and Infinity-8B, respectively. Across different sparsity levels, HACK consistently maintains high generation quality, even under extreme compression (\emph{e.g.}, $\rho=90\%$). The generated images preserve structural layout and semantic consistency despite substantial memory reduction.
Figure\,\ref{fig:2b_sup} and Figure\,\ref{fig:var_sup} further compare HACK with baseline methods (H2O~\cite{zhang2024h2o}, SnapKV~\cite{li2024snapkv}, and LOOK-M~\cite{wan2024look}) on both Infinity-2B and VAR-d30 across multiple compression ratios $\rho$. While baselines degrade at high compression ratios due to the loss of structural information, HACK's head-aware design selectively preserves structural heads. Consequently, HACK outperforms baselines in retaining scene layout and object integrity, demonstrating superior robustness.

\subsection{Visualizations for Attention Patterns}
\begin{figure*}[t]
  \centering
  \includegraphics[width=\textwidth, height=\textheight, keepaspectratio]{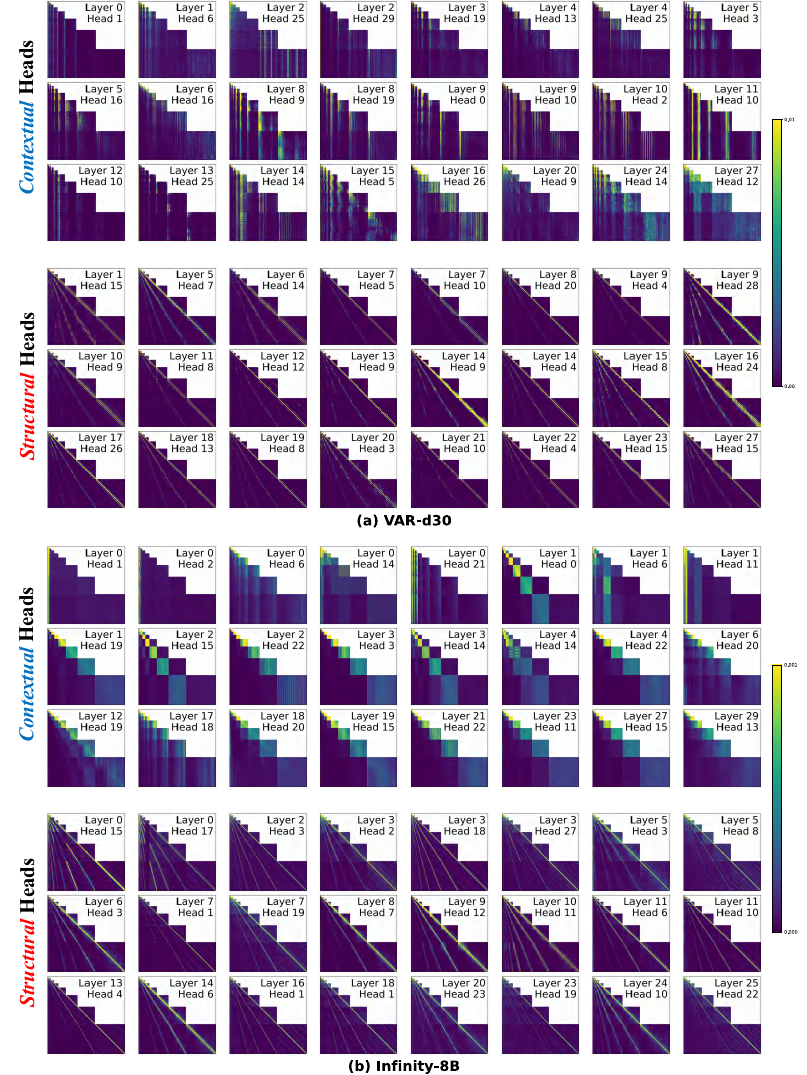} 
  \caption{Additional visualizations of contextual and structural heads in VAR-d30 and Infinity-8B. We concatenate attention maps across scales to facilitate unified visualization.}
  \label{fig:head1}
\end{figure*}

\begin{figure*}[t]
  \centering
  \includegraphics[width=\textwidth, height=\textheight, keepaspectratio]{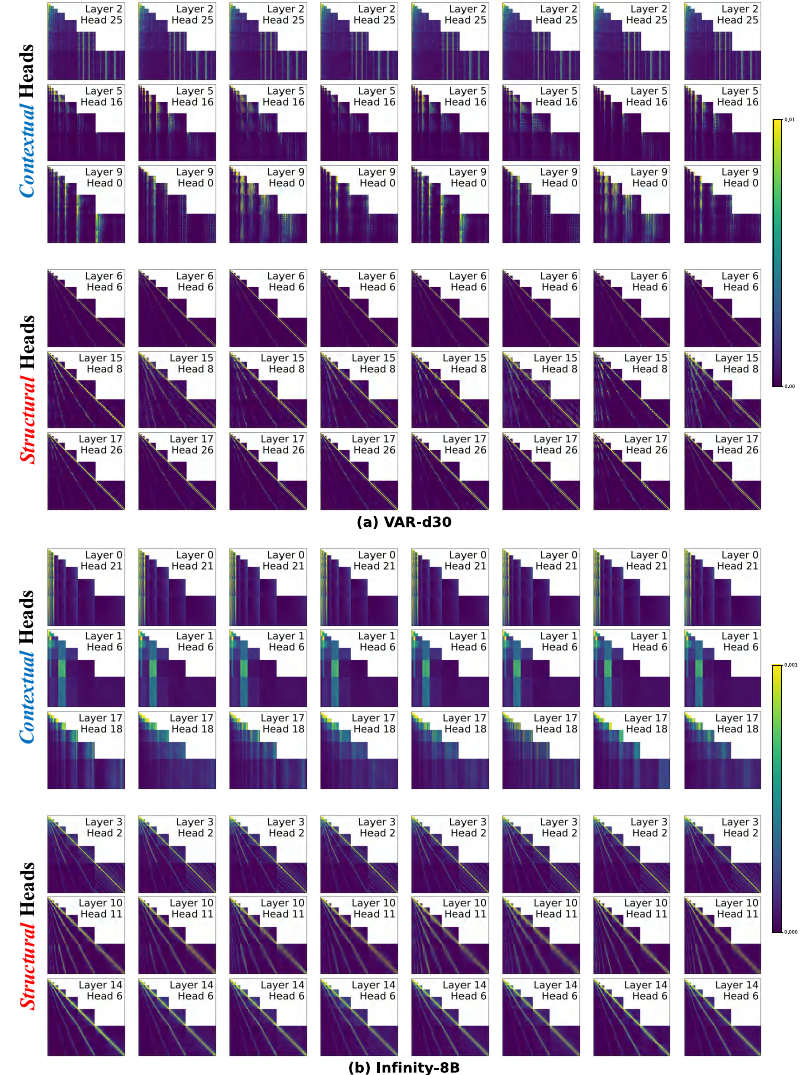} 
  \caption{Additional visualizations of contextual and structural heads across different input samples (columns) in VAR-d30 and Infinity-8B. We concatenate attention maps across scales to facilitate unified visualization.}
  \label{fig:head2}
\end{figure*}

We provide additional attention visualizations to corroborate our analysis of VAR-based models. Figure\,\ref{fig:head1} highlights a distinct dichotomy in Infinity-8B and VAR-d30: \textit{contextual heads} exhibit vertical, global patterns, whereas \textit{structural heads} manifest as diagonal, local dependencies. Moreover, Figure\,\ref{fig:head2} reveals that these patterns are input-agnostic, remaining stable across diverse samples. This confirms that head functionality is an intrinsic property of the model rather than being dependent on specific prompts or labels. Such stability validates the reliability and data efficiency of our offline head classification scheme, solidifying the foundation for our head-aware KV cache compression strategy.

\section{Detailed Complexity Analysis}
\label{proof}

In this section, we provide a detailed derivation of the time complexity for both the vanilla VAR model and the VAR model equipped with HACK. 
For analytical convenience, we assume that each generation step operates on a square resolution, i.e., $h_k = w_k = n_k$, where $k$ denotes the generation scale.
We further assume that the resolution grows geometrically across scales, such that $n_k = a^{k-1}$ for some constant scaling factor $a > 1$. 
Accordingly, the resolution at the final scale $K$ is $n_K = n = a^{K-1}$.

\noindent\textbf{Vanilla VAR.}
In the standard VAR model, each generation step $k$ attends to all previous steps $i \leq k$, and the number of key-value (KV) pairs grows cumulatively across scales. For each scale $k$, the KV cache length $T_k$, can be expressed as summation of tokens number $N_k=n_k^2$:
\begin{equation}
    T_k=\sum_{i=1}^k N_i=\sum_{i=1}^k a^{2(i-1)}=\frac{a^{2k} - 1}{a^2 - 1}.
\end{equation}

The total attention complexity across all steps can be expressed as:

\begin{align}
\small
\sum_{k=1}^{K}N_k\cdot T_k
&= \sum_{k=1}^{K}  a^{2(k - 1)}\cdot \frac{a^{2k} - 1}{a^2 - 1}\\
&= \frac{1-a^{2K}-a^{2K+2}+a^{4K+2}}{(a^{2}-1)^{2}(a^{2}+1)}
\left.\right|_{K=\log_a^{(n)} + 1} \\
&\sim \mathcal{O}( n^4).
\end{align}

This quadratic attention complexity and heavy KV cache rise from accumulated tokens across scales, causing a serious challenge for practical deployment.

\vspace{0.5em}
\noindent\textbf{VAR with HACK.}
HACK addresses this inefficiency by compressing the KV cache at each step, maintaining an average budget within $B$ for the number of cached KV entries. 
As a result, the number of KV pairs involved in attention computation at each step is upper bounded by $B$, and the total attention complexity becomes:
\begin{align}
\small
\sum_{k=1}^{K} N_k \cdot\min \left(T_k, B \right)
&= \sum_{k=1}^{K} a^{2(k-1)}\cdot  \min \left( \frac{a^{2k} - 1}{a^2 - 1}, B \right) \\
&\leq B\cdot\sum_{k=1}^{K} a^{2(k - 1)} \\
&= B \cdot\frac{a^{2K} - 1}{a^2 - 1}\left.\right|_{K=\log_a^{(n)} + 1} \\
&\sim \mathcal{O}( Bn^2).
\end{align}

The additional overhead in HACK arises from KV cache compression, which is triggered only when the total cache length exceeds the budget $B$. Let $k_s$ denote the first step where compression is applied. The main cost comes from the subset attention used to guide this compression. In this step, a small sample of $N_{\text{obs}} \ll N_k$ queries estimates token importance by attending to a key set formed by concatenating the new KV cache from the current step k (length $N_k$) with the previously maintained KV cache (length $B$). The formula for this overhead is:
\begin{equation}
\small
\sum_{k = k_s}^{K} N_{\text{obs}} \cdot (B + N_k) 
= N_{\text{obs}} \sum_{k = k_s}^{K} (B + a^{2(k-1)}) 
\sim \mathcal{O}(N_{\text{obs}} n^2).
\end{equation}
Since $N_{\text{obs}}$ is a small constant (\emph{e.g.}, 32), the compression overhead is negligible in practice.

In summary, HACK reduces the attention complexity of VAR models from $\mathcal{O}(n^4)$ to $\mathcal{O}(Bn^2)$, while incurring only a negligible additional cost of $\mathcal{O}(N_{\text{obs}} n^2)$ due to compression. This validates HACK as an efficient and scalable solution for autoregressive generation.

\section{Limitations and Future Work}
Despite HACK's effectiveness, our work primarily addresses memory and computational bottlenecks within attention modules. A promising avenue for future work is to explore synergies with orthogonal efficiency techniques. For instance, integrating collaborative decoding~\cite{guo2025fastvar} could help mitigate the high costs of large-scale generation. Furthermore, incorporating quantization~\cite{xie2024litevar} and token pruning~\cite{guo2025fastvar} to reduce Feed-Forward Network (FFN) latency could yield additional improvements in overall VAR efficiency.
Finally, while this work focuses on next-scale generation in VAR models, our insights into the functional roles of attention heads are likely applicable to conventional autoregressive visual generation~\cite{sun2024autoregressive,liu2024lumina-mgpt,wu2024janus,qin2025autoregressive}. We anticipate that the proposed head-aware strategies offer a generalizable perspective that can inform the design of efficient KV cache management across a broader spectrum of autoregressive visual architectures.

\end{document}